\documentclass[10pt,twocolumn,letterpaper]{article}

\usepackage[pagenumbers]{cvpr} %

\usepackage{dsfont}
\newcommand{\ind}{\mathds{1}}
\usepackage{booktabs}
\usepackage{makecell}

\usepackage{multirow}
\usepackage{colortbl}     %
\usepackage{xcolor}       %

\definecolor{cvprblue}{rgb}{0.21,0.49,0.74}
\usepackage[pagebackref,breaklinks,colorlinks,allcolors=cvprblue]{hyperref}

\def\paperID{} %
\def\confName{CVPR}
\def\confYear{2026}

\title{CamDirector: Towards Long-Term Coherent Video Trajectory Editing}

\author{
    Zhihao Shi\textsuperscript{1*}, \quad
    Kejia Yin\textsuperscript{2*}, \quad
    Weilin Wan\textsuperscript{3}, \quad
    Yuhongze Zhou\textsuperscript{4}, \quad
    Yuanhao Yu\textsuperscript{1}, \quad \\
    Xinxin Zuo\textsuperscript{5}, \quad
    Qiang Sun\textsuperscript{2,6}, \quad
    Juwei Lu\textsuperscript{2} \\
    \textsuperscript{1}McMaster University \quad
    \textsuperscript{2}University of Toronto \quad
    \textsuperscript{3}The University of Hong Kong \\
    \textsuperscript{4}McGill University \quad
    \textsuperscript{5}Concordia University \quad
    \textsuperscript{6}MBZUAI \quad
}

\begin{document}
\twocolumn[{%
  \renewcommand\twocolumn[1][]{#1}%
  \maketitle
  \vspace{-8mm}
  \begin{center}
    \includegraphics[width=\linewidth]{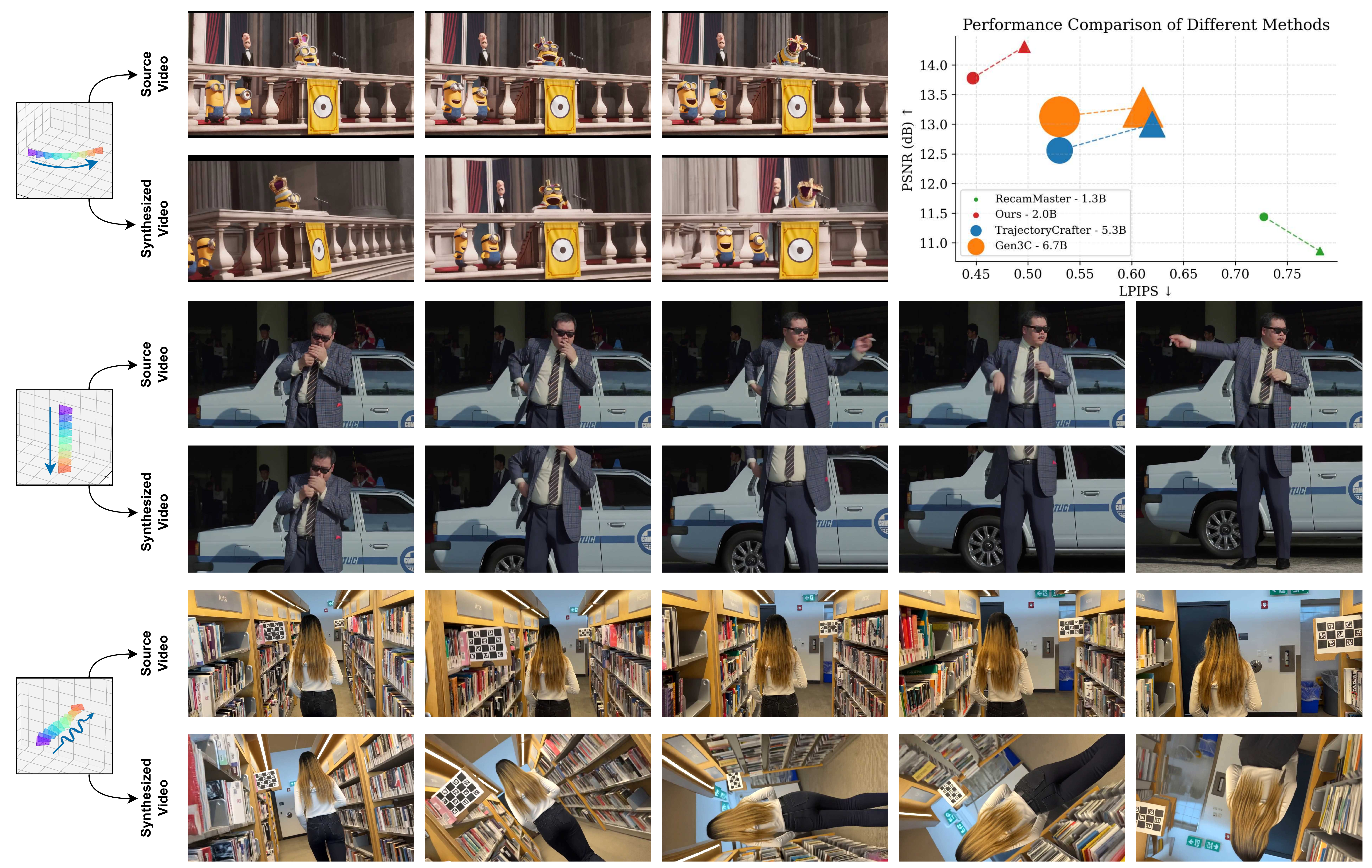}
    \captionof{figure}{
      We present a novel video trajectory editing framework capable of generating new videos along desired trajectories from the given ones to achieve aesthetically pleasing and cinematic camera movements. The proposed method performs favorably against state-of-the-art methods with notably fewer parameters ($\bullet$ and $\blacktriangle$ distinguish different benchmarks). Additional results and videos are available on our \href{https://yinkejia.github.io/CamDirector-Project-Page}{project page}.
    }
    \label{fig:teaser}
  \end{center}
  \vspace{4mm}
}]
\def\thefootnote{*}\footnotetext{Equal contribution.}

\begin{abstract}

Video (camera) trajectory editing aims to synthesize new videos that follow user-defined camera paths while preserving scene content and plausibly inpainting previously unseen regions, upgrading amateur footage into professionally styled videos. 
Existing VTE methods struggle with precise camera control and long-range consistency because they either inject target poses through a limited-capacity embedding or rely on single-frame warping with only implicit cross-frame aggregation in video diffusion models. 
To address these issues, we introduce a new VTE framework that 
1) explicitly aggregates information across the entire source video via a hybrid warping scheme. Specifically, static regions are progressively fused into a world cache then rendered to target camera poses, while dynamic regions are directly warped; their fusion yields globally consistent coarse frames that guide refinement.
2) processes video segments jointly with their history via a history-guided autoregressive diffusion model, while the world cache is incrementally updated to reinforce already inpainted content, enabling long-term temporal coherence. 
Finally, we present iPhone-PTZ, a new VTE benchmark with diverse camera motions and large trajectory variations, and achieve state-of-the-art performance with fewer parameters.

\end{abstract}
    
\section{Introduction}
\label{sec:intro}

Video (camera) trajectory editing (VTE) aims to synthesize new videos from given ones by following newly designed camera trajectories while preserving the original scene content and plausibly inpainting unseen regions.
This technique enables the transformation of amateur videos, often constrained by limited hardware and user expertise, into professionally styled videos characterized by aesthetic composition and cinematic camera motions, effectively bridging the gap between casual users and professional creators.

Leveraging pre-trained video diffusion models, \cite{van2024generative, bai2025recammaster} conditioned the generation process on the input source video and injected target camera poses directly via embedding layers. However, due to the embedding layers' limited representational capacity, these methods lack precise camera control, the generated video cannot follow the target camera trajectory. To achieve better trajectory controllability, \cite{yu2025trajectorycrafter, ren2025gen3c} explicitly warped the source frames along the target trajectory, which were then refined by video diffusion models with missing regions being inpainted.
These methods construct each warped frame from only a single source frame, relying on the bidirectional attention mechanism to implicitly gather complementary information from other frames to ensure consistency.
However, for longer videos, memory constraints require chunked processing, which precludes attention over the full sequence and, in particular, over future frames. Consequently, two key challenges arise: 
1) source alignment: content generated in the current frame may misalign with scene evidence present elsewhere in the source video;
2) self-consistency: the newly inpainted or previously unseen regions can drift across segments, producing temporal flicker and incoherent appearance over time.

To address these challenges, we propose a new video trajectory editing framework with two key components:
1) a hybrid warping scheme that leverages a world cache to aggregate information from the entire source video when constructing each coarse frame, providing a global reference that ensures strong alignment with the original scene content; and
2) a history-guided autoregressive model together with a progressive world-cache update that ensures later segments remain aligned with earlier ones for stable and coherent results.

\begin{figure}[t]
    \setlength{\tabcolsep}{2pt}  %
    \centering
    \begin{tabular}{ccc}
        \includegraphics[width=0.15\textwidth]{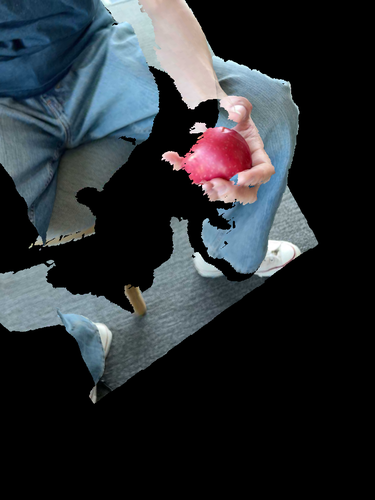} &
        \includegraphics[width=0.15\textwidth]{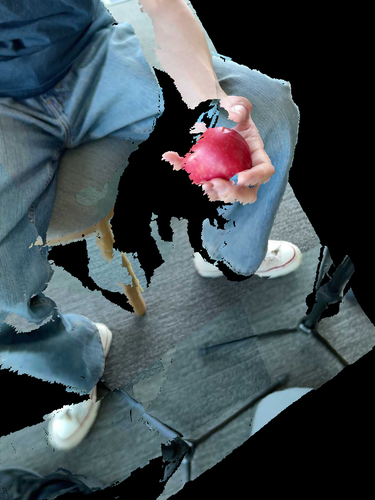} &
        \includegraphics[width=0.15\textwidth]{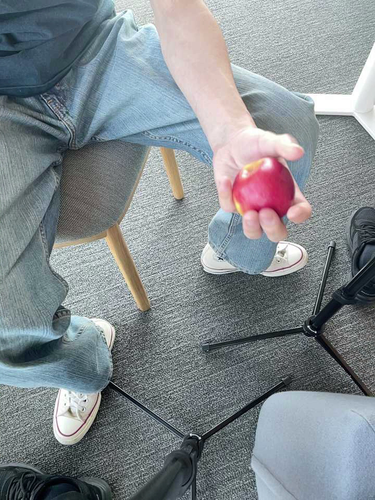} \\
        (a) & (b) & (c)
    \end{tabular}
    \vspace{-3mm}
    \caption{Visual comparison between per-frame warping (a), our hybrid warping (b), and ground truth (c). Hybrid warping tends to produce more complete and source-aligned coarse frames.}
    \vspace{-3mm}
    \label{fig:per_frame-ours-gt}
\end{figure}

Specifically, first, in the proposed hybrid warping, we explicitly decouple the scene into dynamic and static regions. The dynamic areas are directly warped to the target viewpoint to preserve motion fidelity, while static regions are progressively aggregated to construct an unified point cloud, termed the world cache, which is then rendered to the target camera poses. The warped dynamic content and rendered static content are fused to form coarse frames that exhibit higher completeness and consistency, as illustrated in Fig.~\ref{fig:per_frame-ours-gt}.
This hybrid warping efficiently leverages all source frames, including those far apart, to produce globally consistent results. Second, to enforce self-consistency, we introduce a coarse-video–controlled diffusion model (CCDM) that generates short segments and scales to long videos via a history-guided autoregressive scheme. More specifically, in each denoising iteration, two consecutive segments (history and current) are processed together: the cleaner history segment guides the denoising of the current segment. After each segment, we expand the world cache with newly recovered static regions, so subsequent coarse videos encode a stable scene structure and reinforce already inpainted content.
Altogether, seamless inter-segment transitions and long-term visual coherence are effectively ensured.

Finally, since the commonly used iPhone dataset \cite{gao2022monocular} is limited to only five usable test scenes, simple camera motions, small source–target trajectory differences, and narrow fields of view, we introduce a new dataset of ten diverse scenes, called \textbf{iPhone-PTZ}. It spans rotation, dolly, pan, and orbiting motions; includes much larger trajectory variations; covers varied scene content; and provides substantially wider FOVs, yielding a more comprehensive and challenging benchmark for evaluating and advancing VTE methods.

\noindent In summary, our contributions are as follows:
\begin{enumerate}
    \item We propose a novel video trajectory editing framework capable of generating source-aligned and temporally consistent outputs, achieving state-of-the-art performance with fewer parameters as shown in Fig.~\ref{fig:teaser}.

    \item Two novel components are proposed, a Hybrid Warping Scheme that produces complete and well-aligned coarse frames, and a History-Guided Autoregressive Model with progressive world-cache updates, jointly ensuring visually pleasing results with strong source alignment, temporal smoothness, and long-term coherence.

    \item We build a new and challenging VTE benchmark featuring diverse camera motions, large trajectory variations, and rich scene contents, facilitating comprehensive evaluation and future research in video trajectory editing.
\end{enumerate}

\section{Related Work}
\subsection{Video Trajectory Editing}
Given an input video, video trajectory editing aims at generating a new video along target trajectory while preserving the original scene content.
Existing methods \cite{ren2025gen3c, bai2025recammaster, yu2025trajectorycrafter, van2024generative, cao2025uni3c, huang2025voyager, yu2024viewcrafter, ma2025follow} can be broadly divided into two categories.
The first directly conditions video generation on the source video and input target camera poses through embedding layers.
GCD \cite{van2024generative} decomposes relative camera extrinsics into rotation and translation and injects them into the network through an MLP. RecamMaster \cite{bai2025recammaster} encodes target camera extrinsics into a latent space and fuses them with video features. 
However, these methods struggle to precisely control output trajectories, especially in scenarios outside the training distribution where the true metric scale is unknown.
The second category adopts a \textit{warp-and-repainting} paradigm by leveraging explicit 3D cues.
TrajectoryCrafter \cite{yu2025trajectorycrafter} and Gen3C \cite{ren2025gen3c} warp source frames into the target viewpoints using reprojected 3D point clouds, then synthesize the final output via a video diffusion model conditioned on geometrically accurate coarse frames.
Although \cite{yu2025trajectorycrafter, ren2025gen3c} enable more precise camera control, they construct each coarse frame from only a single source frame and rely on bidirectional attention to implicitly gather complementary information from other frames, which becomes problematic in long-video scenarios.

\subsection{Long Video Generation}

Existing VTE methods either fail or exhibit limited performance in generating long videos, primarily due to the absence of mechanisms for maintaining long-term temporal consistency.
Several recent studies have explored various mechanisms for achieving consistent long video generation in T2V and I2V settings, including but not limited to keyframes-to-video~\cite{blattmann2023stable, zhou2024storydiffusion, zhang2023i2vgen}, high compression~\cite{zhang2025packing, jin2024pyramidal, hacohen2024ltx}, discrete temporal chunks~\cite{dobrosotskaya2000magi, gao2024ca2, chen2023seine}, forcing-based autoregressive~\cite{chen2024diffusion, song2025historyguidedvideodiffusion, yin2025slow}.
However, how to apply these strategies to the VTE domain remains an open yet challenging problem, as VTE not only requires temporal self-consistency within the generated video but also demands strict spatial alignment with the source video.
We address this in this work through combining two complementary strategies: a hybrid warping scheme that ensures source alignment and an autoregressive paradigm that promotes temporal coherence.

\section{Method}

\begin{figure*}[t]
    \centering
    \includegraphics[width=\textwidth]{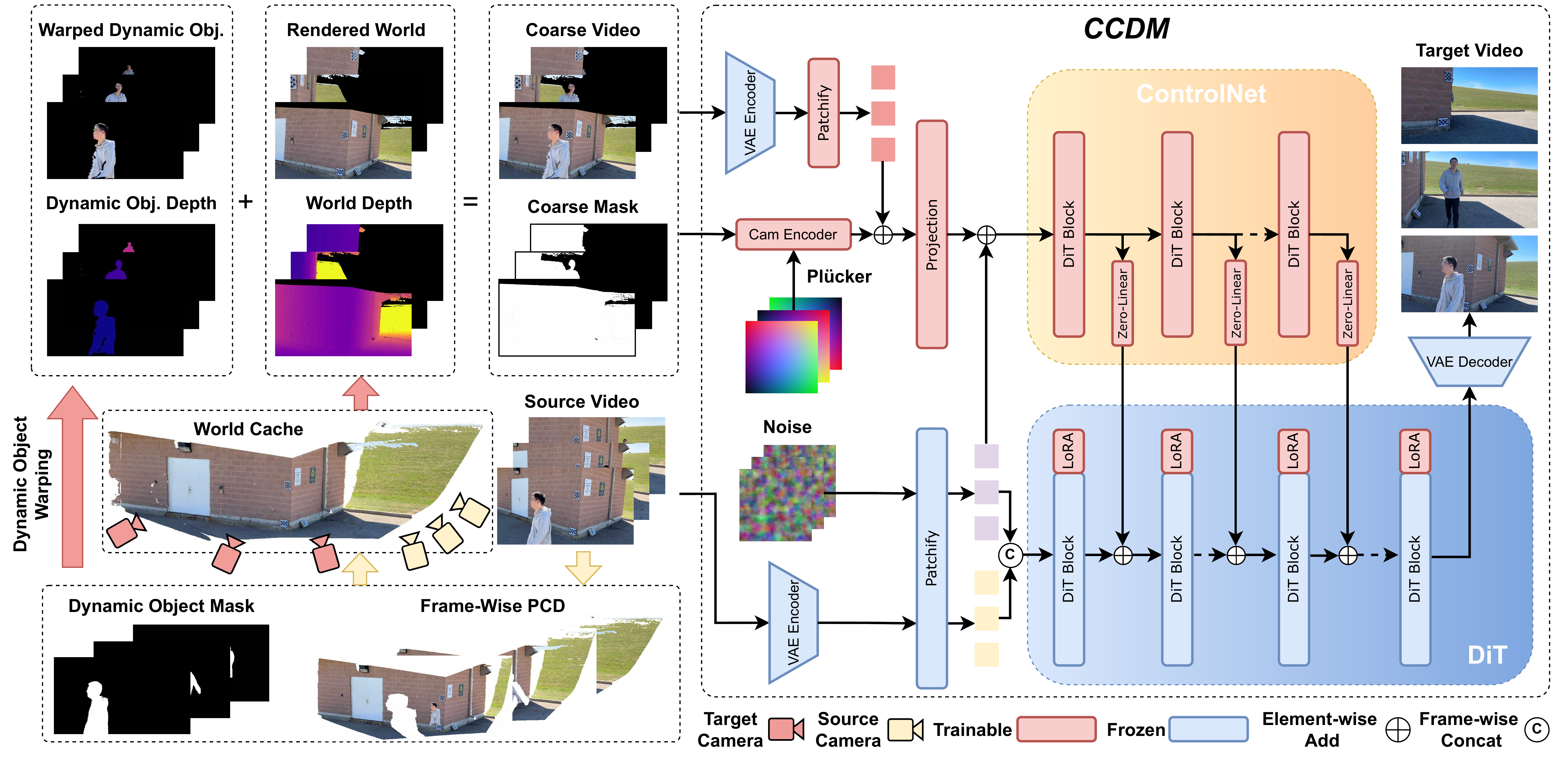}
    \caption{\textbf{Overview of our framework}. \textbf{Left:} The hybrid warping scheme leverages the entire source video to construct coarse frames by processing dynamic and static regions separately, providing a global reference of the original scene content.
    \textbf{Right:} The CCDM conditions the generation on the coarse video via ControlNet, while source-frame tokens are concatenated with target tokens as inputs to the base T2V model to provide reliable motion and appearance priors.}
    \vspace{-3mm}
    \label{fig:main}
\end{figure*}

In this section, we present the proposed framework in detail. As shown in Fig.~\ref{fig:main}, it consists of two main components. First, given a source video, a hybrid warping module synthesizes a coarse video along the target camera trajectory while enforcing 3D and temporal consistency (Sec.~\ref{sec:hybrid}). Second, we design a CCDM to generate videos while leveraging the guidance from both the source video and the rendered coarse video; a novel history-guided autoregressive scheme is proposed to scale the generation to longer videos (Sec.~\ref{sec:vdm}).

\subsection{Hybrid Warping Scheme}
\label{sec:hybrid}
Given a source video $I^s = \{I_i^s\}^N_{i=1} \in \mathbb{R}^{N\times H \times W \times 3}$, we first employ a 4D foundation model, Pi3 \cite{wang2025pi} to estimate point clouds (PCD) for each frame $P = \{P_i\}^N_{i=1} \in \mathbb{R}^{N\times H \times W \times3} $ and their corresponding camera poses $\{\Pi^s_i\}^N_{i=1} $. Considering that dynamic objects evolve over time while static backgrounds should be consistent over the entire scene, we propose a robust and efficient hybrid warping scheme that processes dynamic and static regions differently. 

First of all, we segment the dynamic objects in each frame using a motion segmentation method \cite{huang2025segment} to get the dynamic object masks $\{M^d_i\}^N_{i=1}$. Then, we warp the dynamic regions from each source frame to the target view in a one-to-one manner to preserve motion fidelity. Specifically, each pixel in the mask has its associated 3D point and corresponding color value, and it is warped to the target viewpoints as follows:
\begin{equation}
    I_i^{d,t}, Z_i^{d, t}, M_i^{d,t} = \Phi(\Pi^t_i \cdot (\Pi^s_i)^{-1} \cdot ([P_i, I_i^s] \odot M^d_i)),
\end{equation}
where $\Phi$ is the perspective projection, $I_i^{d,t}$, $I_i^{d,t}$, and $M_i^{d,t}$ represent the warped RGB image, the projected depth, and the valid warped regions, respectively. 

On the other hand, static regions remain consistent across the video, allowing us to construct a unified 3D world cache representation.
A straightforward solution is to naively fuse all static-region PCDs. However, this leads to prohibitive memory usage and computational cost when $N$ is large.
To address this, we propose to construct a lightweight world cache that eliminates redundant points while preserving the original geometric layout.
Specifically, we uniformly sample $L$ out of $N$ frames and iteratively render the world cache at each sampled frame to obtain a visibility mask, the PCDs corresponding to static regions outside the mask are incrementally appended to the cache. 
After traversing all $L$ frames, we obtain a compact yet complete world cache, which substantially reduces memory consumption and improves efficiency.

Finally, we render the world cache at each target viewpoint to produce rendered RGB image $I_i^{w,t}$ and rendered depth $Z_i^{w, t}$, which are then fused with the dynamic counterparts to form the coarse frame: 
$\hat{I_i}(x) = I_i^{d,t}(x) \cdot \ind \!\left(Z_i^{d,t}(x) < Z_i^{w,t}(x)\right)
             + I_i^{w,t}(x) \cdot \ind \!\left(Z_i^{d,t}(x) \ge Z_i^{w,t}(x)\right)$.
Additionally, a visibility mask $M_i^{w,t}$ is generated during rendering, which is merged with $M_i^{d,t}$ to obtain the valid coarse mask $\hat {M_i}$.

The proposed scheme acts as an explicit mechanism capable of perceiving the entire video, not limited by the memory constraints as prior works. 
The produced coarse videos exhibit higher completeness and stronger temporal consistency, substantially reducing the regions requiring inpainting and improving controllability and consistency.

\subsection{History-Guided AutoRegressive Generation}
\label{sec:vdm}
Leveraging the above coarse video, a coarse-video–controlled diffusion model (CCDM) is first designed as a base model to generate video in short segments, which is further adapted to support long video outputs via a novel history-guided autoregressive generation. 

\vspace{1mm}
\noindent \textbf{Base model:}
Fig.~\ref{fig:main} illustrates the overall architecture of the CCDM.
It is built upon a pre-trained video diffusion model, Wan-T2V-1.3B~\cite{wan2025}, for its strong performance and efficiency. 
To preserve generalization ability and photo-realistic synthesis, we incorporate coarse video guidance via ControlNet~\cite{zhang2023adding}, along with its corresponding masks, to help the model distinguish regions to be inpainted.
Following \cite{cao2025uni3c, bahmani2025ac3d}, we also include the target camera pose, encoded as Plücker embeddings, to further improve the visual quality of the generated video. Control features are injected into the first 15 blocks of Wan-T2V, since camera information is primarily determined in the shallow blocks of video diffusion models~\cite{bahmani2025ac3d, liang2025wonderland}.

\begin{figure}[t]
    \centering
    \includegraphics[width=0.48\textwidth]{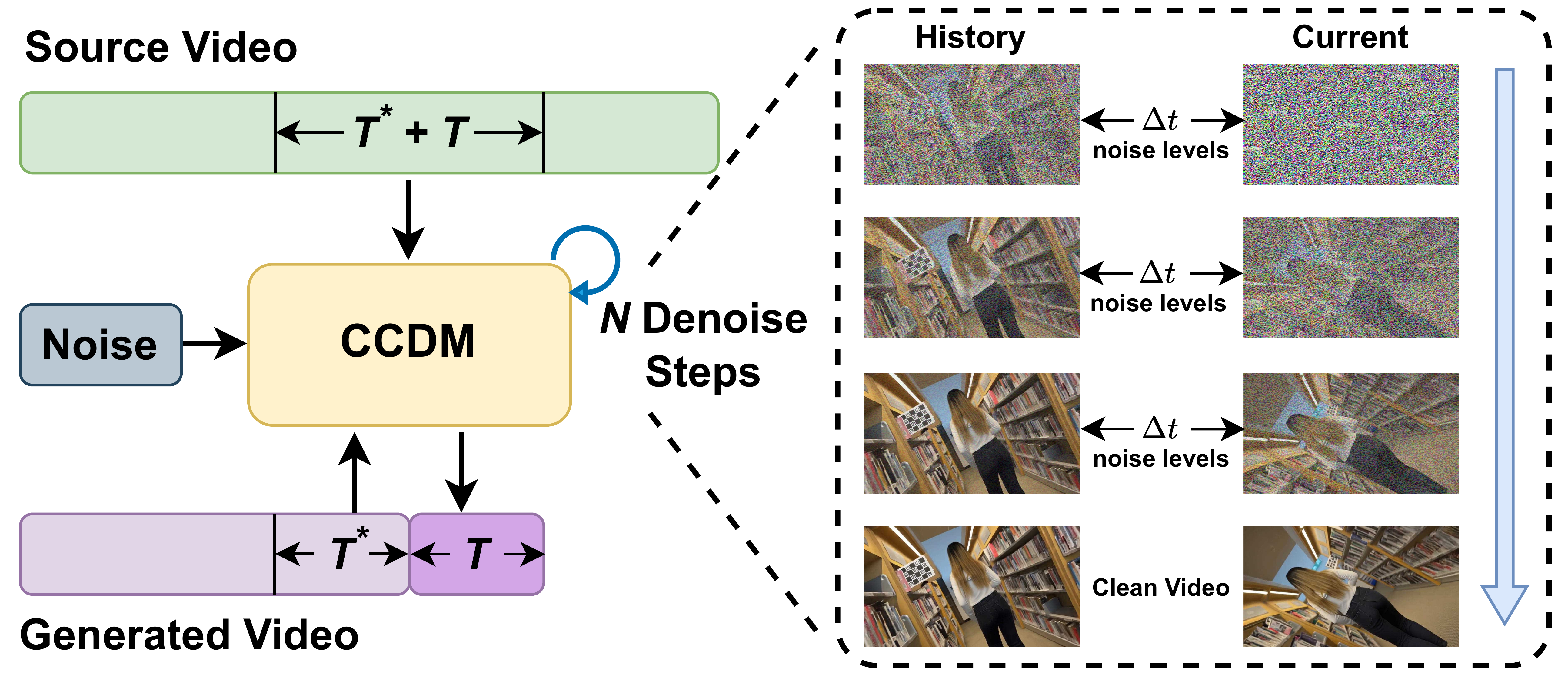}
    \caption{\textbf{Illustration of history-guided autoregressive generation}. 
    In each iteration, $T^{\star}$ previously generated frames serve as history to guide the synthesis of the next $T$ frames, along with the corresponding $T^{\star}+T$ source frames as input to produce the coarse frames and provide original scene context.}
    \vspace{-3mm}
    \label{fig:autoregressive}
\end{figure}

However, the coarse video may contain structural distortions and appearance mismatches due to errors in pose/PCD estimation and view-dependent effects. Therefore, it requires refinement beyond the simple inpainting of unseen regions. To address this, we concatenate source-frame tokens with noisy target tokens, enabling the model to leverage reliable motion and appearance priors through a full attention layer~\cite{bai2025recammaster}.
To accommodate the source latents, we integrate LoRA layers into the original attention modules as an effective and efficient means of adaptation.

\vspace{1mm}
\noindent \textbf{Autoregressive generation:}
Taking the CCDM as a base model, we develop a history-guided, segment-wise autoregressive generation strategy for consistent long-video generation.
As shown in Fig.~\ref{fig:autoregressive}, the long video generation process is divided into a sequential generation of multiple non-overlapping segments $\{x_k\}^K_{k=1}$, each containing $T$ consecutive frames.
In each iteration, the last $T^{\star}$ frames of the previously generated segment are used as history to guide the synthesis of the current segment of $T$ frames.
Specifically, the history tokens and the tokens in the current segment jointly form the target noisy tokens in CCDM, allowing history context information to be propagated across video segments through the attention layers.
Empirically, we find that maintaining the history tokens $\Delta t$ noise steps ahead of the current tokens throughout the denoising process yields the most consistent results. After denoising, the current clean segment is then re-corrupted to the corresponding noise levels of the subsequent segment to serve as the history progressively.
To further strengthen the guidance and achieve a smooth transition, we introduce the
classifier-free guidance~\cite{ho2022classifier}. 
The predicted flow at each denoising step is computed as:
\begin{equation}
    v_t = w \times v_{\theta}(x^{k}_{t-1} | x^{k-1}_{t+\Delta t}) + (1 - w) \times v_{\theta}(x^{k}_{t-1} | x^{k-1}_{t-1}),
\end{equation}
where $v_{\theta}$ and $w$ denote our network and the guidance scale, respectively. 

\begin{figure}[t]
    \centering
    \includegraphics[width=0.45 \textwidth]{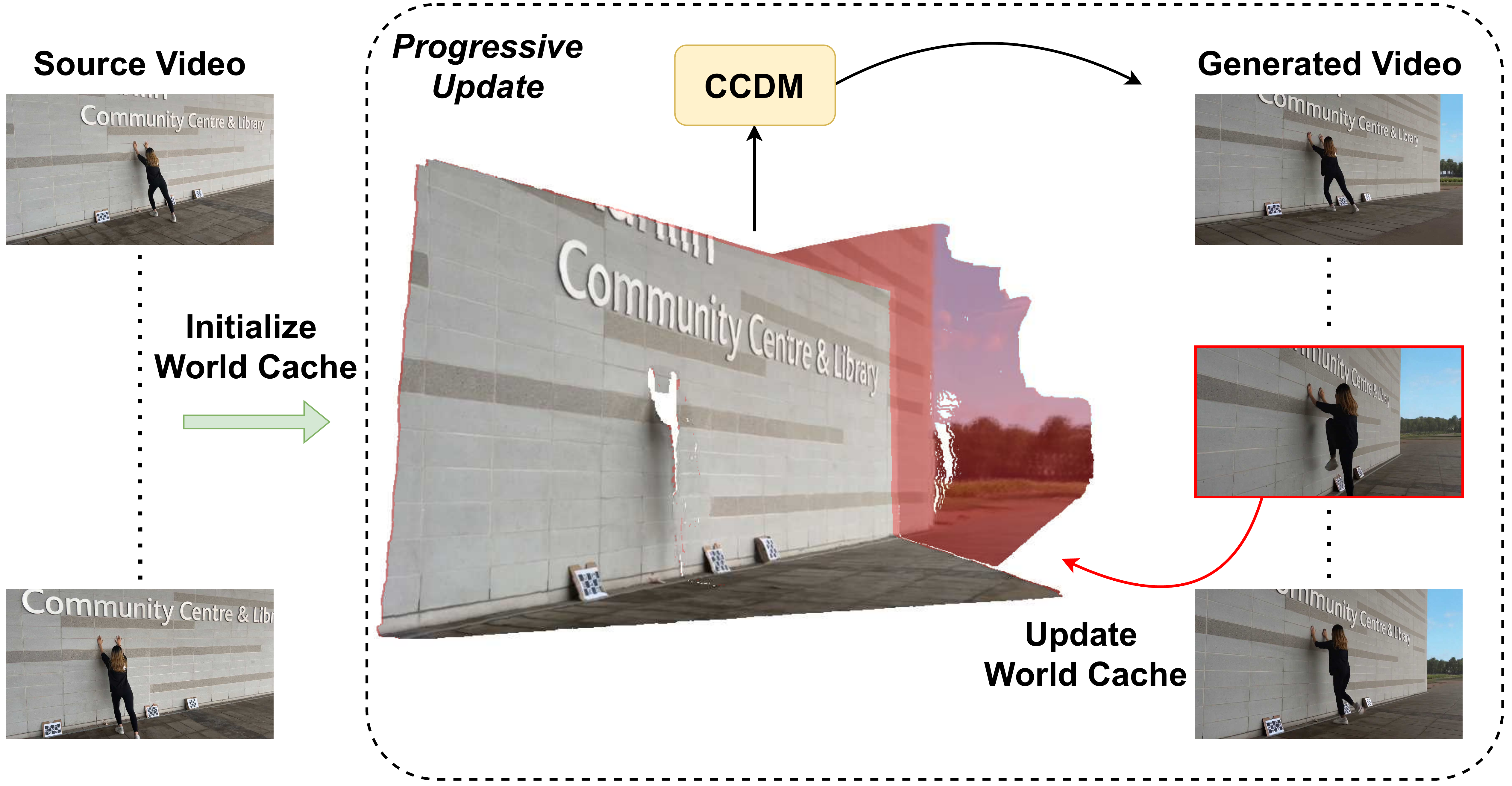}
    \vspace{-3mm}
    \caption{\textbf{Illustration of progressive world cache update}. 
    Whenever a new segment is generated, we evenly sample $C$ frames as anchors, where the newly inpainted regions are merged into the world cache. The updated regions are highlighted in red.}
    \vspace{-3mm}
    \label{fig:dynamic_update}
\end{figure}

Finally, to further enhance long-term coherence, we progressively update the world cache when new segments are generated (see Fig.~\ref{fig:dynamic_update}).
First, we employ SAM2 \cite{ravi2024sam} to trace the static regions in the source video segment and the newly synthesized segment. Then we use Pi3~\cite{wang2025pi} to estimate its point cloud, which is subsequently aligned to the world coordinate. Afterward, we merge the newly inpainted regions into the existing world cache (see the supplementary materials for details).
As such, the coarse video produced for subsequent segments incorporates the newly inpainted static regions, allowing later segments to align more effectively with earlier ones, and together with history guidance, ensures seamless transitions across segments and long-term temporal consistency in the generated video.

\begin{figure}[t]
    \centering
    \setlength{\tabcolsep}{2pt}
    \begin{tabular}{cccc}
        \includegraphics[width=0.11\textwidth]{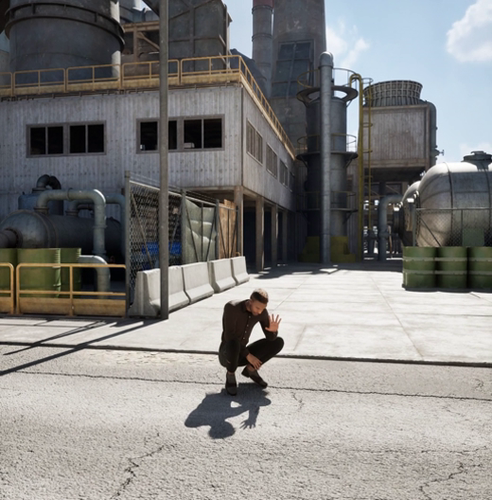} &
        \includegraphics[width=0.11\textwidth]{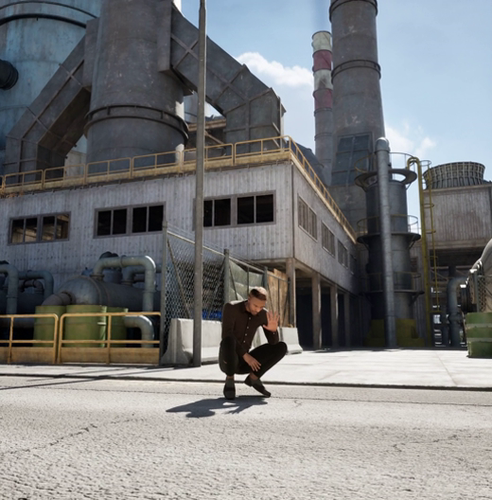} &
        \includegraphics[width=0.11\textwidth]{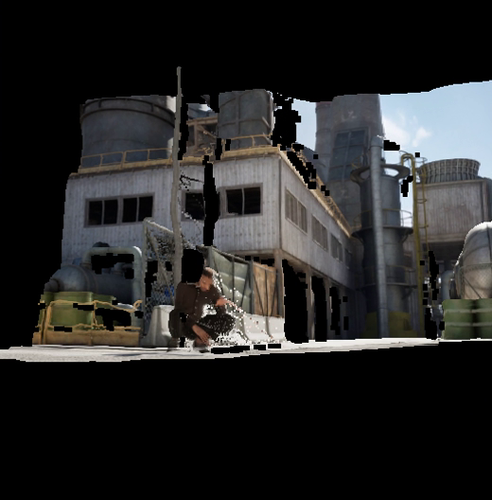} &
        \includegraphics[width=0.11\textwidth]{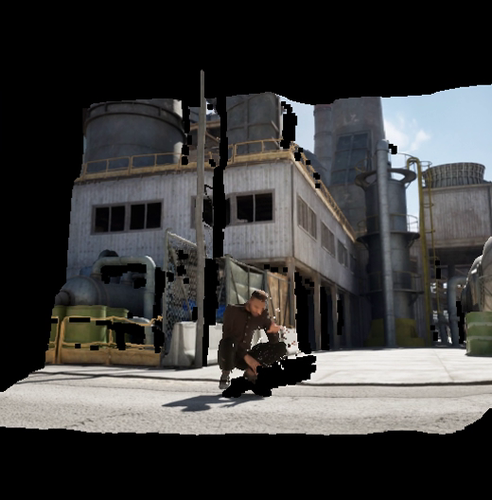} \\
        Source &  Target &  VGGT &  Aligned
    \end{tabular}
    \vspace{-3mm}
    \caption{\textbf{Illustration of VGGT depth vs. our corrected depths}. The last two columns visualize warping results from the source to the target view using VGGT's estimated depth and our corrected depth, respectively.
    VGGT’s depth leads to a poor warping result (third column), whereas our adjusted depth yields a more aligned outcome (last column).}
    \vspace{-3mm}
    \label{fig:vggt_error}
\end{figure}

\vspace{1mm}
\noindent \textbf{Training:}
We adopt a dynamic multi-view dataset~\cite{bai2025recammaster} to train our model, which contains approximately $13.6K$ dynamic scenes, each comprising ten synchronized 81-frame videos with corresponding camera poses.
However, since PCDs and depth maps are not provided, additional data processing is required. 
Specifically, at each synchronized time step, ten frames from the videos form a static multi-view capture, for which we estimate depth and camera poses using VGGT~\cite{wang2025vggt} and align the results to the ground-truth camera coordinate system. 
Nevertheless, we observe that VGGT sometimes produces inaccurate depth maps, resulting in erroneous coarse videos, as illustrated in Fig.~\ref{fig:vggt_error}.  To address these issues, we correct the estimated depths using epipolar constraints and apply a series of filtering rules to remove low-quality samples, such as those with abrupt changes between consecutive frames, resulting in $9.5$K scenes for training.
See supplements for details.

Training is carried out in two stages:
1) First, we train the base CCDM. In each iteration, two videos are randomly sampled as the source and target. A coarse video is generated from the source using our hybrid warping scheme, and the target video is corrupted with noise sampled uniformly between 0 and 1000. The model is trained using the standard flow-matching objective~\cite{lipman2022flow}.
2) Second, we fine-tune CCDM to support autoregressive generation. Similar to the previous training stage, from the source video, a coarse video is generated; for the target video, it is divided into $T^{\star}$ frames as the history and $T$ frames as the current segment. Two non-decreasing noise levels $(t_1 \le t_2)$ are applied to the history and current, respectively, and flow-matching losses are used for both.

\begin{table*}[t]
\centering
\small 
\setlength{\tabcolsep}{3.5pt}  %
\renewcommand{\arraystretch}{1.3}  %
\begin{tabular}{cccccccc}
    \hline
    \multirow{2}{*}{Method} & 
    \multirow{2}{*}{\# Params} & 
    \multicolumn{3}{c}{iPhone \cite{gao2022monocular} } &
    \multicolumn{3}{c}{iPhone-PTZ } \\
    \cmidrule(lr{0.5em}){3-5} \cmidrule(lr{0.5em}){6-8}
     & & PSNR $\uparrow$ & LPIPS $\downarrow$ & FID $\downarrow$
     & PSNR $\uparrow$ & LPIPS $\downarrow$ & FID $\downarrow$ \\
    \hline
    RecamMaster \cite{bai2025recammaster} & 1.3B
                & 10.73 / - & 0.7830 / - & 195.24 / - 
                & 11.64 / - & 0.6981 / - & 117.77 / - \\            
    TrajectoryCrafter \cite{yu2025trajectorycrafter} & 5.3B
                & 13.00 / - & 0.6197 / - & 145.58 / - 
                & 12.56 / - & 0.5303 / - & 105.30 / - \\
    Gen3C \cite{ren2025gen3c} & 6.7B
                & 13.29 / 13.44 & 0.6107 / 0.6066 & 148.76 / 116.91 
                & 13.13 / 13.27 & 0.5305 / 0.5497 & 91.41 / 86.21 \\
    \rowcolor[gray]{0.95}
    Ours & 2.0B
            & \textbf{14.31 / 14.12} & \textbf{0.4952 / 0.5103} & \textbf{114.99 / 107.44}
            & \textbf{13.78 / 13.99} & \textbf{0.4468 / 0.4752} & \textbf{79.65 / 72.33} \\
    \hline
\end{tabular}
\vspace{-3mm}
\caption{\textbf{Quantitative comparison on iPhone and iPhone-PTZ benchmark}. Results for \textit{short clips} are on the left, \textit{full videos} are on the right. The best results are highlighted in \textbf{bold}. }
\label{tab:main_results}
\end{table*}

\begin{table*}[t]
\centering
\setlength{\tabcolsep}{5pt}  %
\renewcommand{\arraystretch}{1.3}  %
\begin{tabular}{ccccccc}
    \hline
    \multirow{2}{*}{Method} & 
    \multicolumn{6}{c}{iPhone\cite{gao2022monocular} / iPhone-PTZ} \\
    \cmidrule(lr{0.5em}){2-7}
    & \makecell{Subject \\ Consistency $\uparrow$}
    & \makecell{Background \\ Consistency $\uparrow$}
    & \makecell{Temporal \\ Flickering $\uparrow$}
    & \makecell{Motion \\ Smoothness $\uparrow$}
    & \makecell{Aesthetic \\ Quality $\uparrow$}
    & \makecell{Imaging\\ Quality $\uparrow$} \\
    \hline
    Gen3C \cite{ren2025gen3c}
                & 0.8510 / 0.8228 & 0.8900 / 0.8597 & 0.9627 / 0.9376
                & 0.9804 / 0.9819 & 0.4194 / 0.4222 & 0.6703 / 0.7304 \\          
    \rowcolor[gray]{0.95}
    Ours
            & \textbf{0.9400 / 0.8574} & \textbf{0.9489 / 0.8816} & \textbf{0.9801 / 0.9478}
            & \textbf{0.9874 / 0.9844} & \textbf{0.4298 / 0.4435} & \textbf{0.7645 / 0.7557} \\
            
    \hline
\end{tabular}
\vspace{-3mm}
\caption{\textbf{VBench comparison on iPhone and iPhone-PTZ benchmark}. Results for iPhone are on the left, iPhone-PTZ are on the right.. The best results are highlighted in \textbf{bold}.
}
\vspace{-3mm}
\label{tab:main_vbench}
\end{table*}

\section{Experiments}

\begin{figure*}[t]
    \centering
    \setlength{\tabcolsep}{2pt}  %
    \setlength{\fboxsep}{0pt}    %
    \setlength{\fboxrule}{0.5pt}
    \renewcommand{\arraystretch}{1}  %
    \begin{tabular}{cccccc}

        \includegraphics[width=0.15\textwidth]{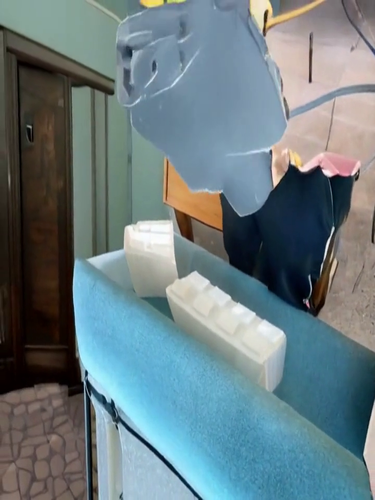} &
        \includegraphics[width=0.15\textwidth]{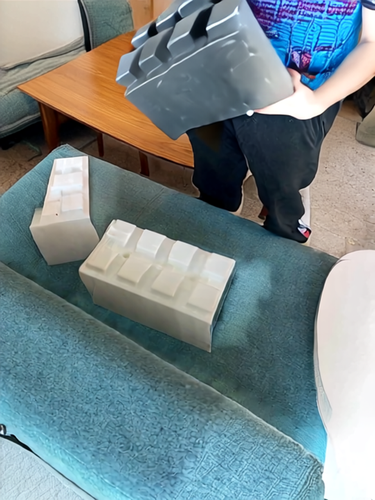} &
        \includegraphics[width=0.15\textwidth]{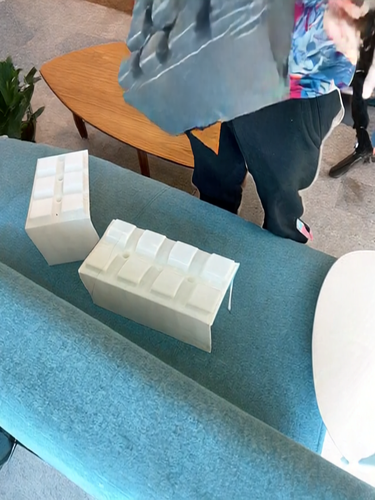} &
        \includegraphics[width=0.15\textwidth]{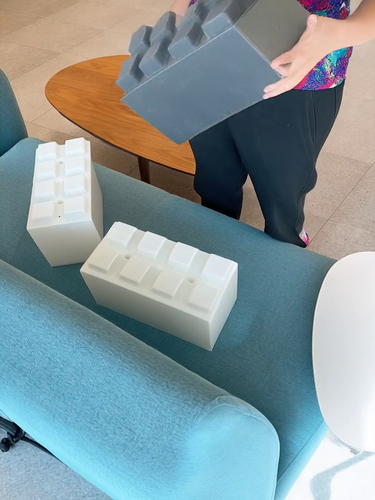} &
        \fbox{\includegraphics[width=0.15\textwidth]{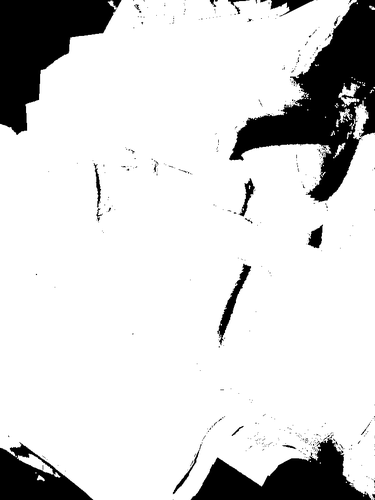}} &
        \includegraphics[width=0.15\textwidth]{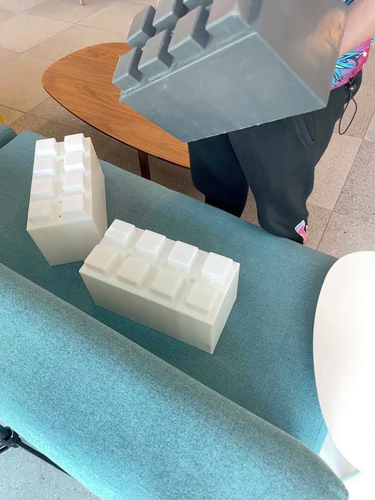} \\

        \includegraphics[width=0.15\textwidth]{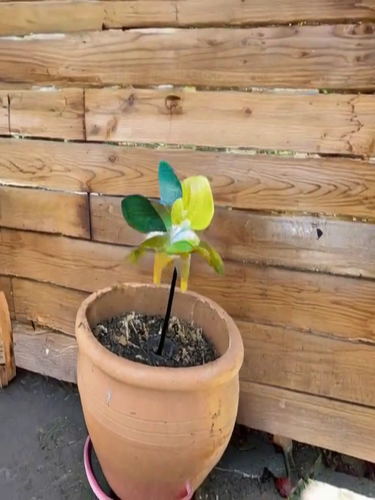} &
        \includegraphics[width=0.15\textwidth]{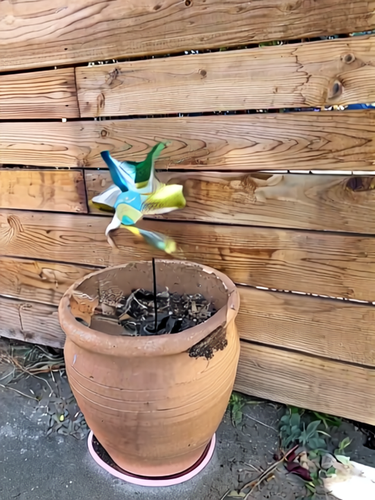} &
        \includegraphics[width=0.15\textwidth]{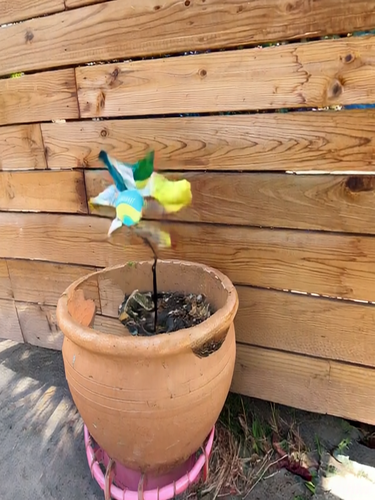} &
        \includegraphics[width=0.15\textwidth]{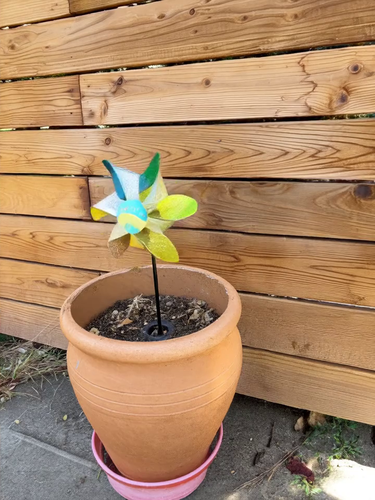} &
        \fbox{\includegraphics[width=0.15\textwidth]{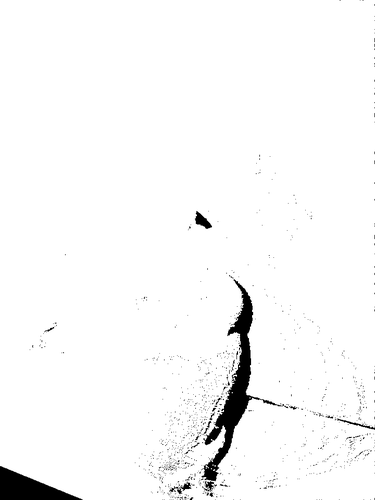}} &
        \includegraphics[width=0.15\textwidth]{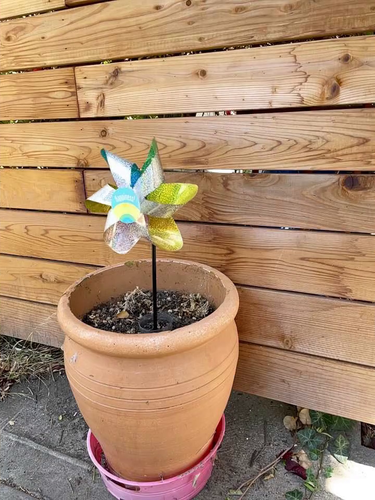} \\

        \includegraphics[width=0.15\textwidth]{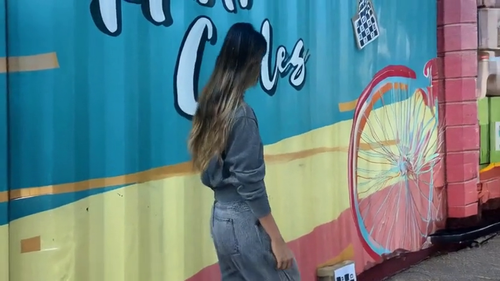} &
        \includegraphics[width=0.15\textwidth]{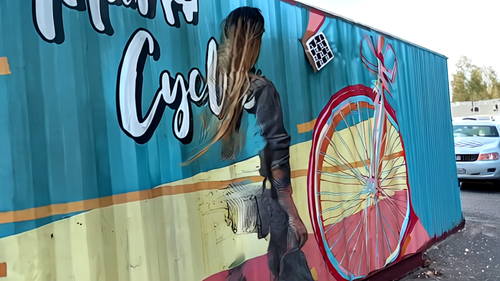} &
        \includegraphics[width=0.15\textwidth]{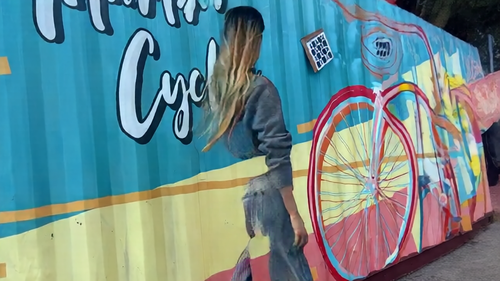} &
        \includegraphics[width=0.15\textwidth]{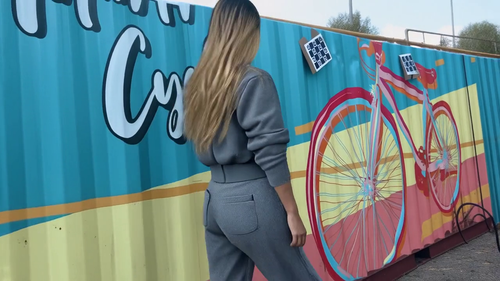} &
        \fbox{\includegraphics[width=0.15\textwidth]{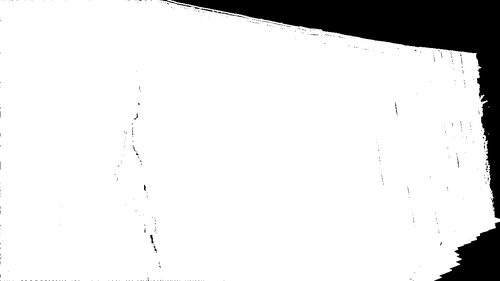}} &
        \includegraphics[width=0.15\textwidth]{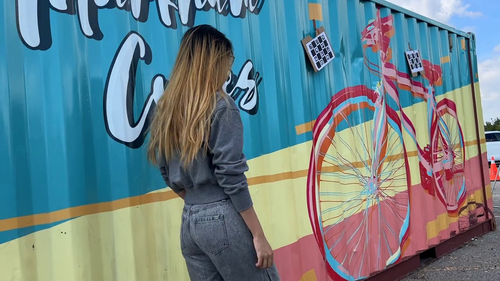} \\

        \includegraphics[width=0.15\textwidth]{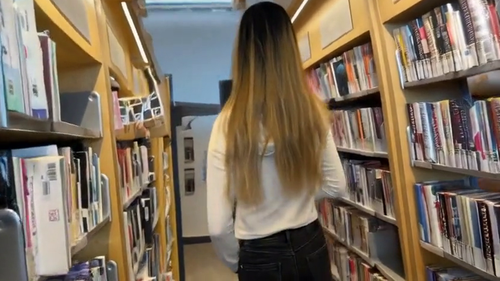} &
        \includegraphics[width=0.15\textwidth]{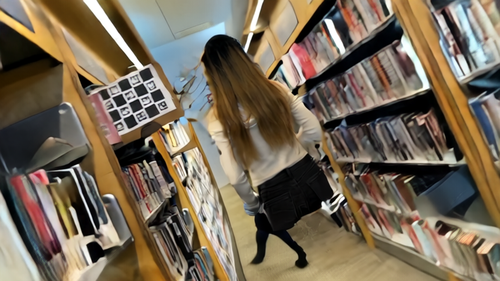} &
        \includegraphics[width=0.15\textwidth]{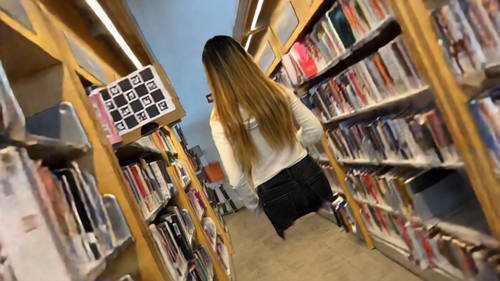} &
        \includegraphics[width=0.15\textwidth]{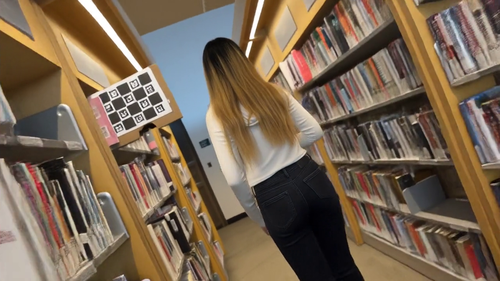} &
        \fbox{\includegraphics[width=0.15\textwidth]{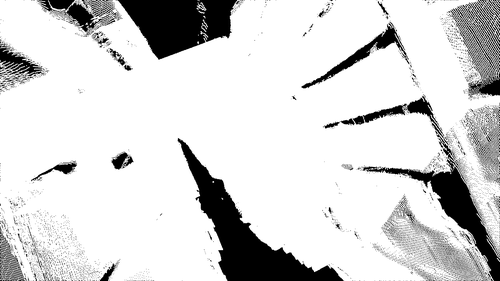}} &
        \includegraphics[width=0.15\textwidth]{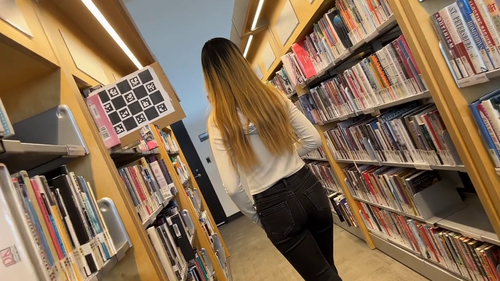} \\

        \includegraphics[width=0.15\textwidth]{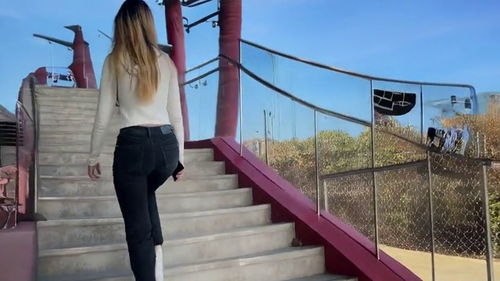} &
        \includegraphics[width=0.15\textwidth]{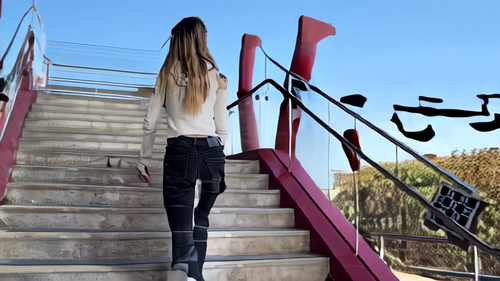} &
        \includegraphics[width=0.15\textwidth]{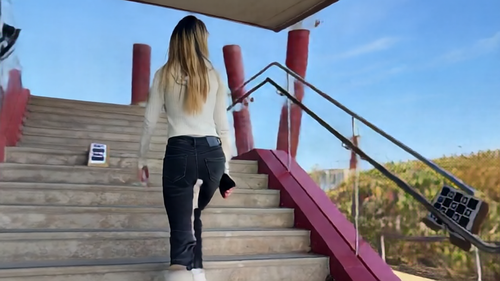} &
        \includegraphics[width=0.15\textwidth]{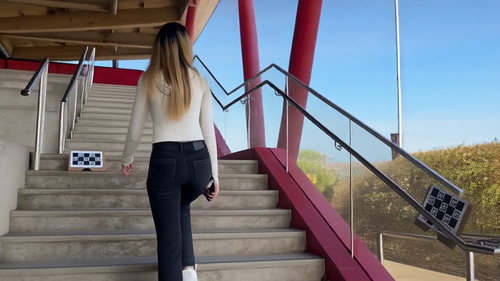} &
        \fbox{\includegraphics[width=0.15\textwidth]{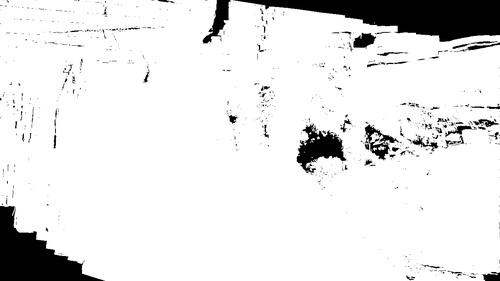}} &
        \includegraphics[width=0.15\textwidth]{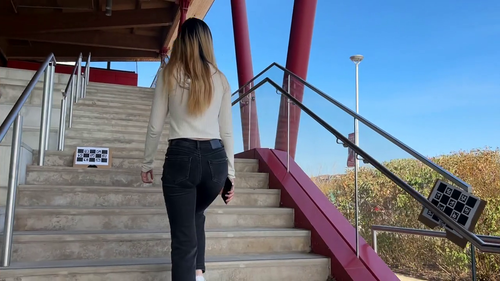} \\
        
        RecamMaster & TrajectoryCrafter & Gen3C & Ours & Visibility Mask & GT 
    \end{tabular}
    \vspace{-3mm}
    \caption{\textbf{Qualitative comparison of short clips on the iPhone and iPhone-PTZ dataset}.}
    \vspace{-2mm}
    \label{fig:qualitative_iphone}
\end{figure*}

\begin{figure*}[t]
    \centering
    \setlength{\tabcolsep}{2pt}  %
    \renewcommand{\arraystretch}{1}  %
    \newlength{\colw}
    \setlength{\colw}{2.9cm}
    \begin{tabular}{m{1cm}*{5}{>{\centering\arraybackslash}m{\colw}}}

        & 2s & 4s & 6s & 8s & $>$ 10s \\
        \centering Gen3C &
        \includegraphics[width=\colw]{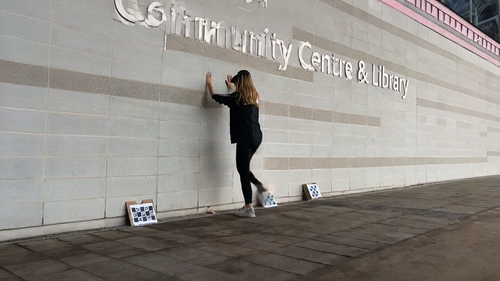} &
        \includegraphics[width=\colw]{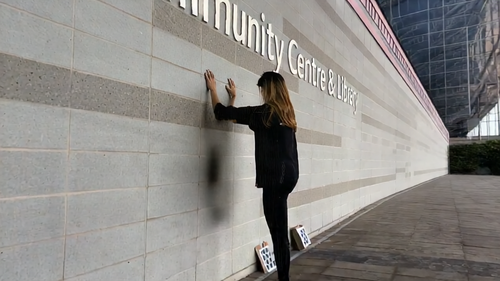} &
        \includegraphics[width=\colw]{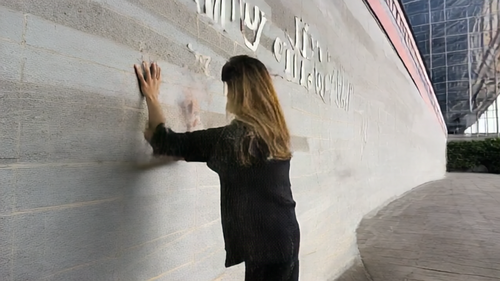} &
        \includegraphics[width=\colw]{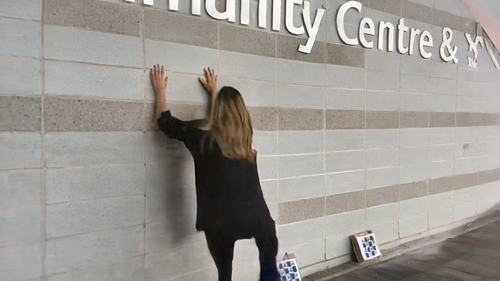} &
        \includegraphics[width=\colw]{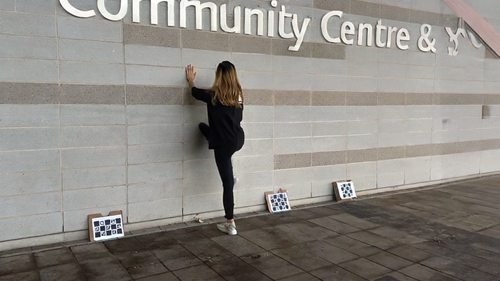} \\

        \centering Ours &
        \includegraphics[width=\colw]{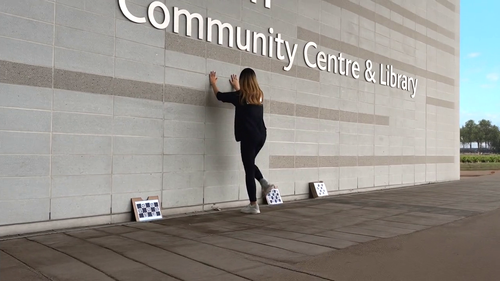} &
        \includegraphics[width=\colw]{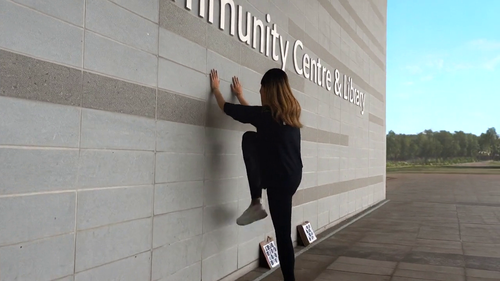} &
        \includegraphics[width=\colw]{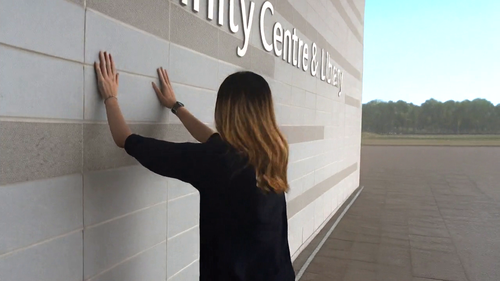} &
        \includegraphics[width=\colw]{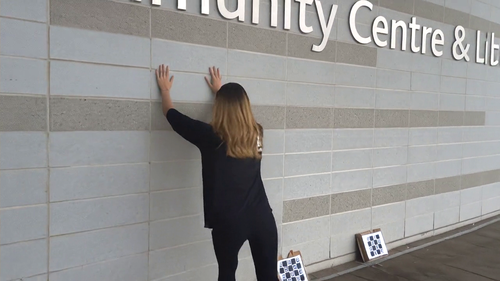} &
        \includegraphics[width=\colw]{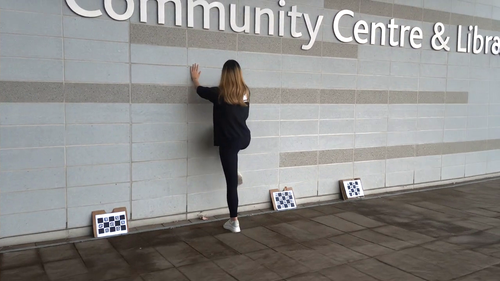} \\

        \centering GT &
        \includegraphics[width=\colw]{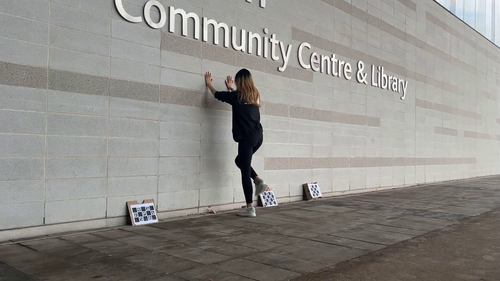} &
        \includegraphics[width=\colw]{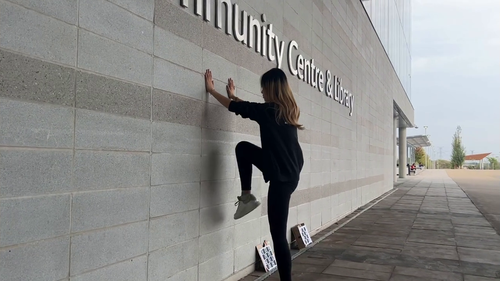} &
        \includegraphics[width=\colw]{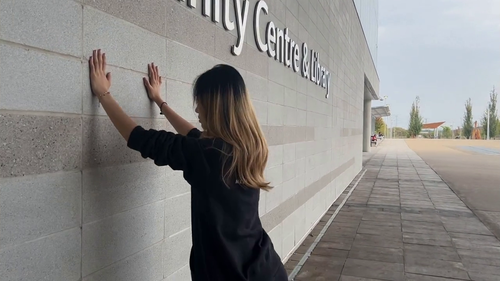} &
        \includegraphics[width=\colw]{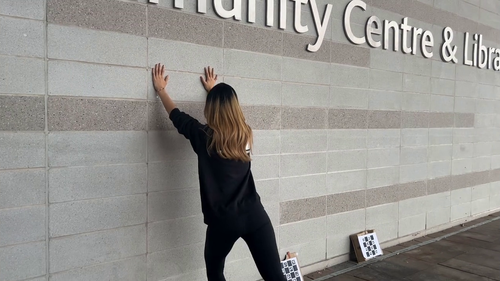} &
        \includegraphics[width=\colw]{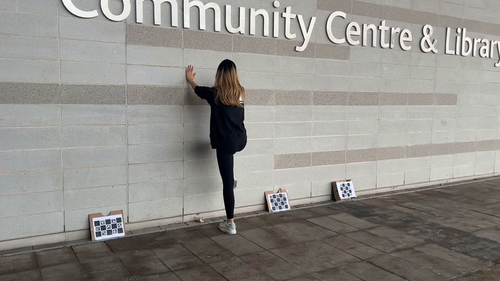} \\

    \end{tabular}
    \vspace{-3mm}
    \caption{\textbf{Qualitative comparison of full videos on iPhone-PTZ dataset}.}
    \label{fig:qualitative_long_video}
\end{figure*}

\subsection{Implementation Details}

We train both the base and autoregressive model on our processed dynamic multi-view dataset (Sec.~\ref{sec:vdm}) for 20,000 steps at a resolution of $480 \times 832$, using a learning rate of $2\times10^{-5}$. 
Training is conducted on six GPUs with a batch size of 6, taking approximately 20 hours for each. The number of target frames is set to 41 for the base CCDM, while $T$ and $T^{\star}$ are set as 20 and 21, respectively, for the autoregressive version.

\subsection{VTE Benchmarks}

\textbf{iPhone \cite{gao2022monocular}.} 
Following prior work~\cite{yu2025trajectorycrafter, wang2025shape}, we use five scenes from the iPhone dataset to evaluate our method, where each scene contains three synchronized videos with a resolution of $720 \times 960$ and duration ranging from 7-15s.
The first video of each scene, captured with casual handheld motion, is treated as the source, while the second video, recorded from a static camera, serves as the target.

\noindent\textbf{iPhone-PTZ.} 
While widely used, the iPhone dataset presents several limitations:
1) limited trajectory diversity, the source and target videos only involve handheld and static captures, failing to represent the wide range of camera motions encountered in real-world scenarios;
2) restricted motion scale, the source videos cover only a small spatial range, resulting in a narrow field of view.
Consequently, the dataset cannot comprehensively evaluate VTE methods, whose key application is to transform diverse casual videos into artistic ones with cinematic camera movements.

To address these limitations, we construct a more suitable and challenging benchmark, termed iPhone-PTZ, which includes ten diverse scenes featuring a broad spectrum of camera motions, such as dolly, pan, orbiting, and significantly larger fields of view.
Each scene contains two synchronized $1280 \times 720$ videos, with duration ranging from 5-12s, captured using identical iPhone 14 Plus devices: the first recorded by casual users under handheld settings, and the second by professional operators using a DJI Osmo Mobile 7P PTZ to introduce cinematic camera motions.

\subsection{Comparison with State-of-the-Art}
We adopt widely used metrics, PSNR, LPIPS \cite{zhang2018perceptual}, and FID \cite{heusel2017gans} to quantitatively compare different methods.
Since RecamMaster and TrajectoryCrafter can only handle short clips, we report results on both short clips (the first 41 frames) and the full video.
As shown in Tab.~\ref{tab:main_results}, our method consistently outperforms prior approaches across all metrics while using significantly fewer parameters, demonstrating the effectiveness of our method.
We further evaluate full-length video quality using VBench~\cite{huang2023vbench} in Tab.~\ref{tab:main_vbench}, where our method again achieves the best performance, particularly in video consistency metrics, highlighting the strength of our long-video modeling design.

In addition, we present qualitative comparisons on two benchmarks in Fig.~\ref{fig:qualitative_iphone}. 
As can be seen from the figures, RecamMaster~\cite{bai2025recammaster} fails to achieve precise camera control, TrajectoryCrafter~\cite{yu2025trajectorycrafter} introduces severe visual artifacts, while Gen3C~\cite{ren2025gen3c} tends to produce incomplete and blurry results.
More importantly, these methods perform poorly in source alignment\footnote{The visibility mask indicates source-visible regions that should match the ground truth, while regions outside the mask are inpainted from scratch.}.
For example, in the second and third cases of Fig.~\ref{fig:qualitative_iphone}, the floor and rear part of the bicycle appear in the other frames of the source videos, yet prior methods fail to exploit this information, leading to misaligned outputs.
In contrast, our method produces spatial-accurate, visual-pleasing, and source-aligned results.

Finally, Fig.~\ref{fig:qualitative_long_video} presents comparisons on full video generation. Gen3C struggles to generate clear and consistent outputs, while our method achieves superior visual fidelity and maintains long-term structural coherence throughout the entire sequence.

\begin{table}[!t]
\centering
\setlength{\tabcolsep}{6pt}  %
\renewcommand{\arraystretch}{1.3}  %
\begin{tabular}{cccc}
    \hline
     & PSNR $\uparrow$ & LPIPS $\downarrow$ & FID $\downarrow$ \\
    \hline
    \rowcolor[gray]{0.95}
    Ours & \textbf{13.99} & \textbf{0.4752} & \textbf{72.33} \\
    \hline
    w/o Plücker & 13.18 & 0.4897 & 78.23 \\
    w/o Source & 13.04 & 0.5134 & 92.90 \\
    w/o Hybrid Warping & 12.18 & 0.5347 & 84.75 \\
    \hline
\end{tabular}
\vspace{-3mm}
\caption{\textbf{Ablation on hybrid warping and CCDM conditions}. }
\label{tab:ablation_short}
\end{table}

\begin{figure}[t]
    \centering
    \includegraphics[width=0.48\textwidth]{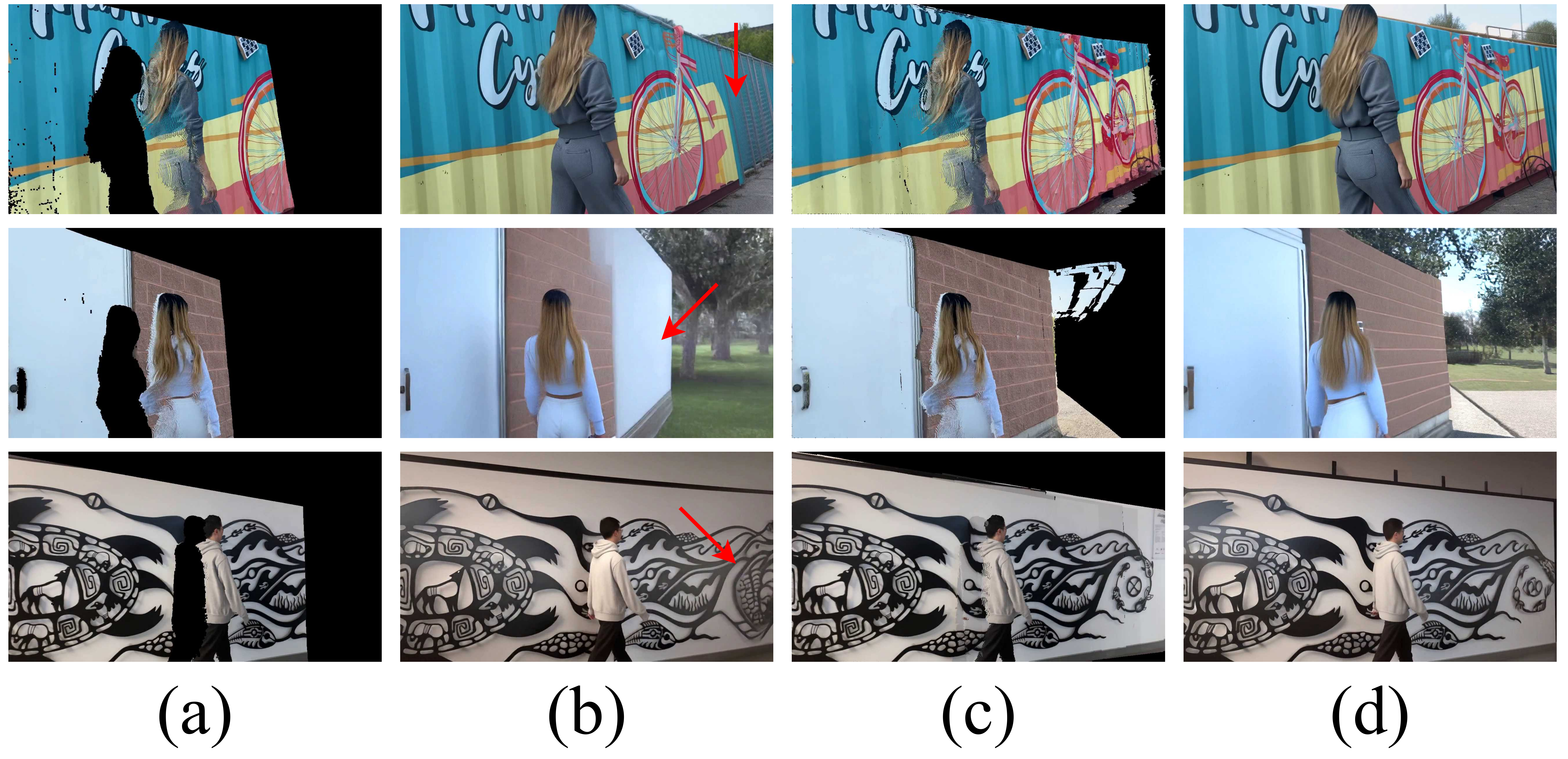}
    \vspace{-6mm}
    \caption{\textbf{Effectiveness of hybrid warping scheme}. We compare per-frame warping (a) and our hybrid warping (c), along with their corresponding final results (b) and (d). Hybrid warping yields more complete and source-aligned coarse frames, leading to higher-quality final generations.}
    \label{fig:ablation_hybrid_warp}
\end{figure}

\subsection{Ablation Study}
In this section, we conduct ablation studies on key components of our proposed method. All experiments are performed in the full-length video setting on the iPhone-PTZ benchmark.

\vspace{1mm}
\noindent \textbf{Ablation on hybrid warping scheme:} 
To examine the impact of the hybrid warping scheme, we replace it with per-frame warping while keeping all other components unchanged. As shown in Tab.~\ref{tab:ablation_short}, this leads to a significant performance degradation, highlighting the importance of hybrid warping. The visual comparison in Fig.~\ref{fig:ablation_hybrid_warp} further confirms this: hybrid warping effectively aggregates complementary information across the entire source video, producing more complete and source-aligned coarse frames, which ultimately lead to higher-quality final generations.

\begin{figure}[t]
    \centering
    \includegraphics[width=0.45\textwidth]{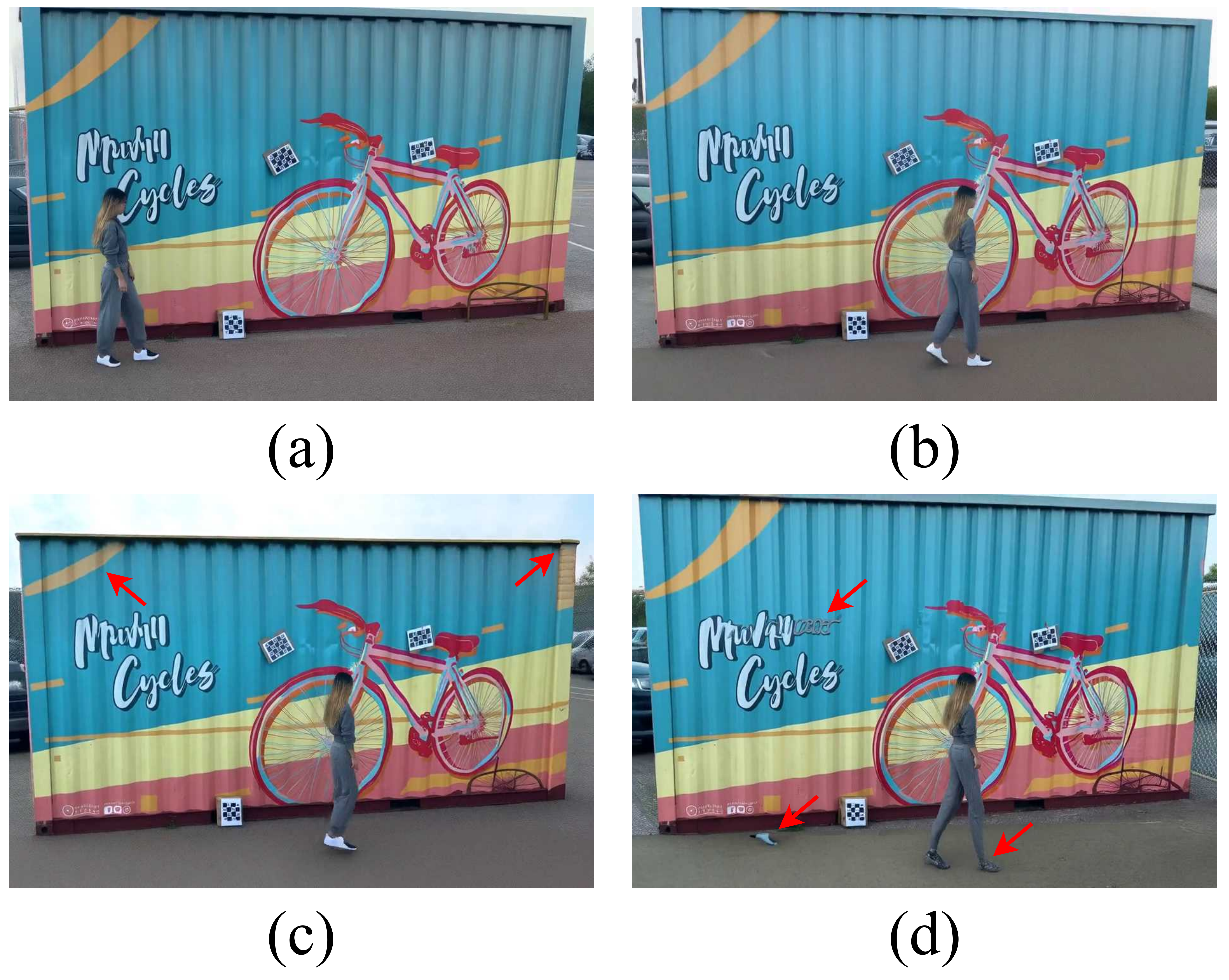}
    \vspace{-3mm}
    \caption{\textbf{Effectiveness of components in the history-guided autoregressive generation}. Starting from the same first frame (a), we compare the last-frame outputs of our full model (b), the variant without progressive world-cache update (c), and the variant without history guidance (d).}
    \label{fig:ablation_long_video}
\end{figure}

\vspace{1mm}
\noindent\textbf{Ablation on history-guided autoregressive generation:}
We investigate the impact of two key components, history guidance and progressive world cache update, by disabling each in turn, with quantitative results reported in Tab.~\ref{tab:ablation_vbench}.
As shown, removing either component leads to performance degradation, demonstrating their individual contributions.
Fig.~\ref{fig:ablation_long_video} further illustrates their complementary roles:
removing the progressive world-cache update prevents later segments from accessing already inpainted content from earlier segments, independent repainting causing inconsistent results as shown in Fig.~\ref{fig:ablation_long_video}(c); 
whereas disabling history guidance results in a drifting scene appearance across segments.
By incorporating both components, our full model achieves coherent and stable results.

\begin{table}[!t]
\centering
\setlength{\tabcolsep}{3pt}  %
\renewcommand{\arraystretch}{1.3}  %
\begin{tabular}{cccc}
    \hline
     & PSNR $\uparrow$ & \small\makecell{Subject \\ Consis.} $\uparrow$ & \small\makecell{Background \\ Consis.} $\uparrow$ \\
    \hline
    \rowcolor[gray]{0.95}
    Ours & \textbf{13.99} & \textbf{0.8574} & \textbf{0.8816} \\
    \hline
    w/o History Guidance & 13.39 & 0.8543 & 0.8780 \\
    w/o Progressive Update & 12.86 & 0.8487 & 0.8777 \\
    \hline
\end{tabular}
\vspace{-3mm}
\caption{\textbf{Ablation on history-guided autoregressive generation}.}
\label{tab:ablation_vbench}
\end{table}

\vspace{1mm}
\noindent\textbf{Ablation on the network conditions.}
Lastly, we ablate the input conditions of CCDM by retraining the model with either the source-video input or the Plücker embedding removed.
As shown in Tab.~\ref{tab:ablation_short}, removing either component leads to a noticeable performance drop, highlighting their contributions to the final output.

\section{Conclusion}
We propose a novel video trajectory editing framework that achieves state-of-the-art performance with significantly fewer parameters. The framework incorporates two complementary components that jointly ensure strong alignment with the source video and robust long-term temporal consistency. Extensive experiments and ablation studies demonstrate the effectiveness of our design.

\vspace{3mm}
\noindent \textbf{Limitations:}
Occasionally, the generated frames appear over-smoothed, especially in regions with complex textures. This is primarily because we train the model on a synthetic dynamic dataset, whose rendered textures are relatively coarse. 
As future work, we plan to incorporate real-world static multi-view datasets as complementary training sources, or construct a new synthetic dataset with richer textures and more vivid details using advanced CG techniques.

{
    \small
    \bibliographystyle{ieeenat_fullname}
    \bibliography{main}
}

\clearpage
\setcounter{page}{1}
\maketitlesupplementary

\section{Dynamic Multi-view Dataset Processing}
We adopt the dynamic multi-view dataset introduced by ~\cite{bai2025recammaster} to train our model. The dataset contains $13.6$K dynamic scenes, each consisting of ten synchronized 81-frame videos with corresponding camera poses. 
To accelerate training, we precompute the dynamic object masks and point clouds for all videos, avoiding repeated calls to the motion-segmentation module and Pi3 during training.

Observing that most videos contain only one dynamic object, a human, we use Grounding DINO~\cite{liu2024grounding} with the text prompt “person” to detect the human’s bounding box in the first frame, and then track it throughout the sequence with SAM2~\cite{ravi2024sam}. The tracked human segmentation masks serve as an effective and efficient approximation of the dynamic object masks.

To generate point clouds for each video with respect to the ground-truth camera poses, we employ VGGT to estimate depth maps and then align them to the ground-truth coordinate system.
Since VGGT is originally designed for static captures, we apply it to the ten synchronized frames at each time step, treating them as a static multi-view observation. The resulting depth maps are then aligned to the ground-truth coordinate system through the following procedure:
1. Keypoint detection and matching: We detect keypoints using SuperPoint~\cite{detone2018superpoint} and match correspondences across the ten views using LightGlue~\cite{lindenberger2023lightglue}.
2. Sparse 3D reconstruction: Using the ground-truth camera poses, we triangulate the matched correspondences to obtain sparse 3D points.
3. Depth alignment: The VGGT-predicted depths are aligned to the sparse 3D points via a linear transformation, performed separately for foreground and background regions.

Finally, we apply a series of filtering steps to remove invalid samples as well as those exhibiting strong inconsistencies or severe artifacts in the coarse videos synthesized using the precomputed dynamic masks and depth maps.
A sample is discarded if any of the following conditions occur:
1. Grounding DINO fails to detect any bounding boxes, leaving no dynamic object mask for constructing coarse videos;
2. the intersection-over-union (IoU) between consecutive coarse-frame masks falls below 0.6, indicating unstable or inconsistent reconstruction.

\section{Progressive World Cache Update}
Given a newly generated segment, the goal is to merge the newly inpainted static regions into the world cache.
To achieve this, we sample 2 frames from the generated segment and concatenate them with the 5 uniformly sampled frames from the corresponding source segment that were used when building the initial world cache.
We then apply SAM2 to trace the static regions and use Pi3 to estimate their point clouds.
We treat the source frames as a bridge to align the estimated point clouds to the world cache coordinate:
a linear transformation between the existing world-cache point clouds and the newly estimated point clouds of the source frames is computed using Umeyama algorithm \cite{umeyama2002least} and applied to the point clouds of the sampled target frames.
Finally, we merge the static portions of the aligned point clouds from the sampled frames into the world cache.

\section{IPhone-PTZ Benchmark Description}
The benchmark consists of 10 scenes, each containing two synchronized 1280 × 720 videos, with duration ranging from 5-12s, captured with identical iPhone 14 Plus devices.
The first video is recorded by casual users under handheld settings, characterized by noticeable hand shake, off-center framing, and non-artistic camera motion.
The second video is recorded by professional operators using a DJI Osmo Mobile 7P PTZ gimbal, introducing cinematic motions such as Inception-style rotations, human tracing, stable orbiting, and other smooth controlled trajectories.

To obtain the camera pose for each frame, we first mask out dynamic objects using SAM2, then run COLMAP on the static regions to estimate camera trajectories. To further improve robustness and reconstruction stability, several calibration boards are placed in the scene during capture.

A preview of the proposed benchmark can be found in our uploaded demo video.

\section{More qualitative results}
\begin{figure*}[t]
    \centering
    \setlength{\tabcolsep}{2pt}  %
    \setlength{\fboxsep}{0pt}    %
    \setlength{\fboxrule}{0.5pt}
    \renewcommand{\arraystretch}{1.1}  %
    \begin{tabular}{cccccc}

        \includegraphics[width=0.15\textwidth]{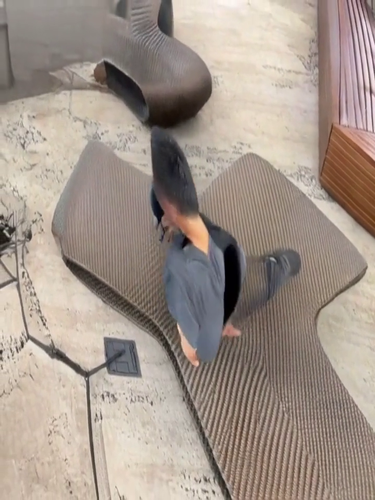} &
        \includegraphics[width=0.15\textwidth]{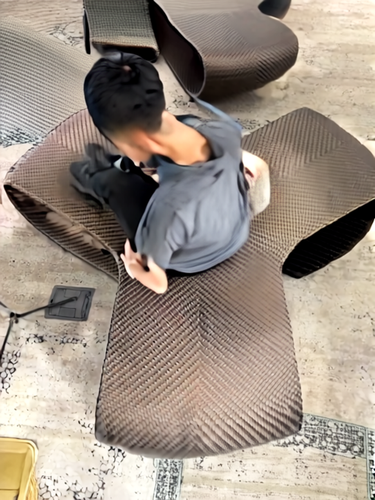} &
        \includegraphics[width=0.15\textwidth]{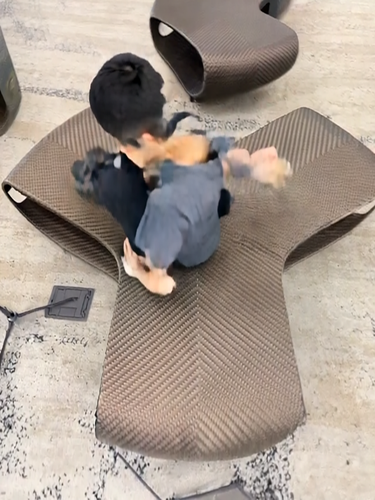} &
        \includegraphics[width=0.15\textwidth]{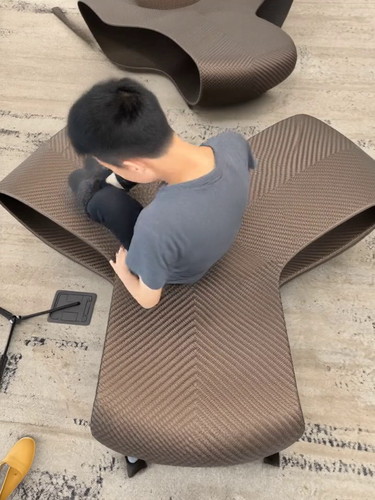} &
        \fbox{\includegraphics[width=0.15\textwidth]{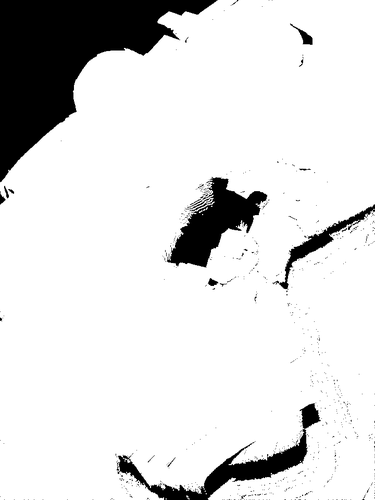}} &
        \includegraphics[width=0.15\textwidth]{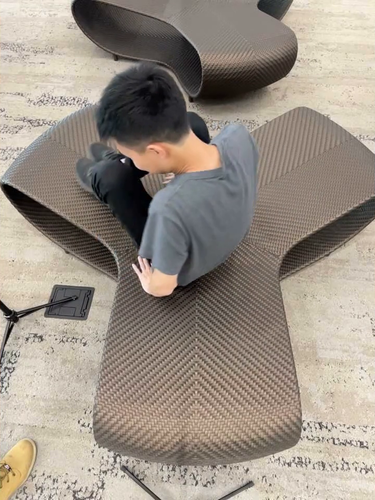} \\

        \includegraphics[width=0.15\textwidth]{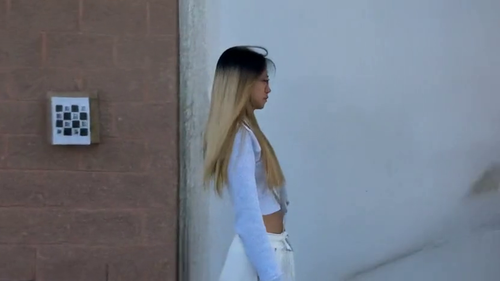} &
        \includegraphics[width=0.15\textwidth]{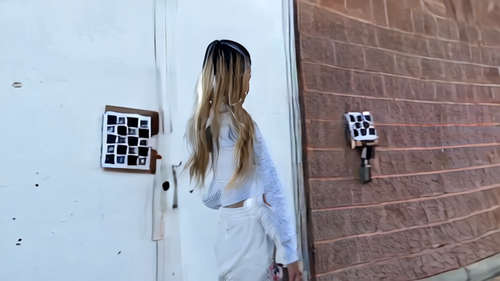} &
        \includegraphics[width=0.15\textwidth]{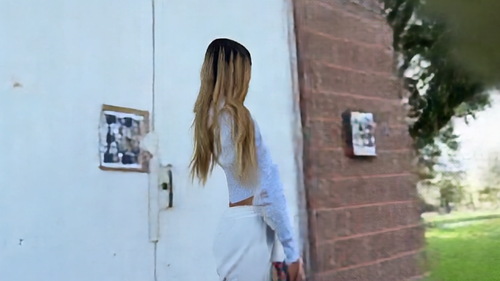} &
        \includegraphics[width=0.15\textwidth]{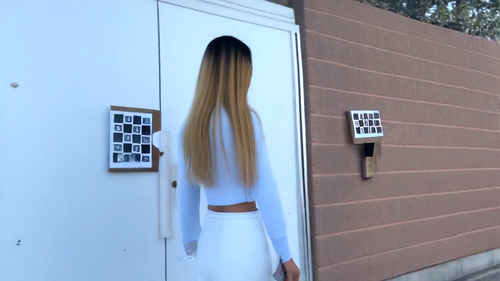} &
        \fbox{\includegraphics[width=0.15\textwidth]{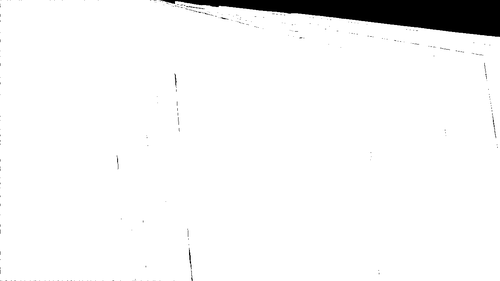}} &
        \includegraphics[width=0.15\textwidth]{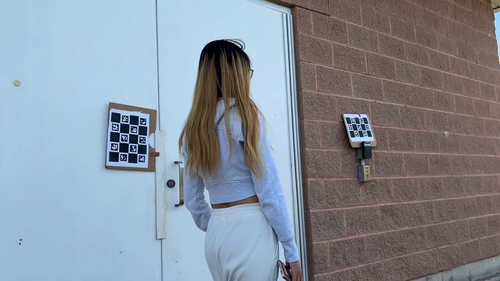} \\

        \includegraphics[width=0.15\textwidth]{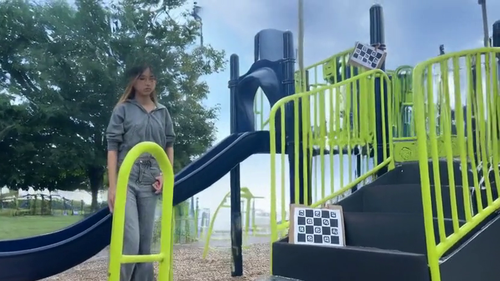} &
        \includegraphics[width=0.15\textwidth]{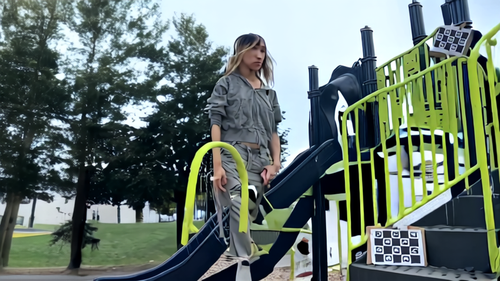} &
        \includegraphics[width=0.15\textwidth]{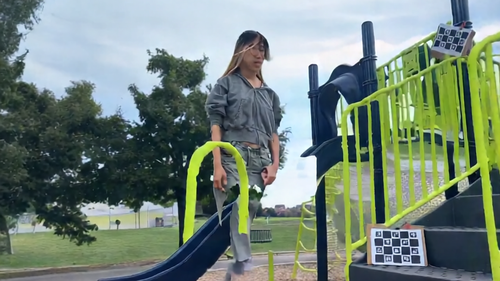} &
        \includegraphics[width=0.15\textwidth]{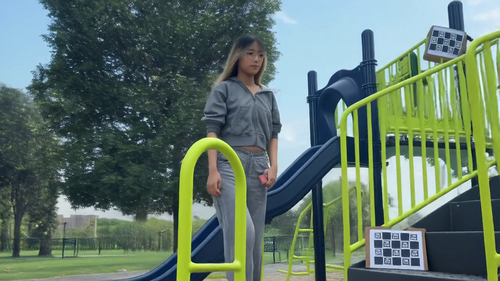} &
        \fbox{\includegraphics[width=0.15\textwidth]{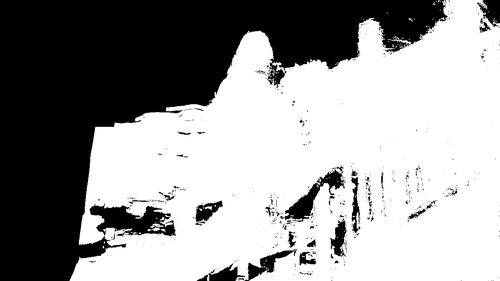}} &
        \includegraphics[width=0.15\textwidth]{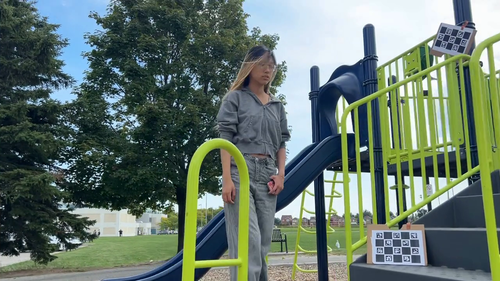} \\

        \includegraphics[width=0.15\textwidth]{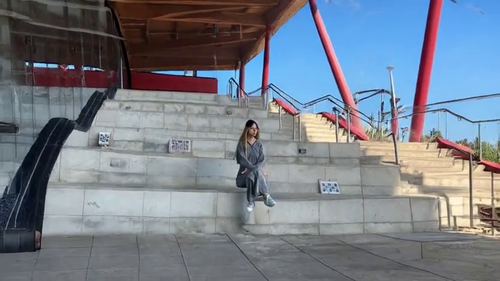} &
        \includegraphics[width=0.15\textwidth]{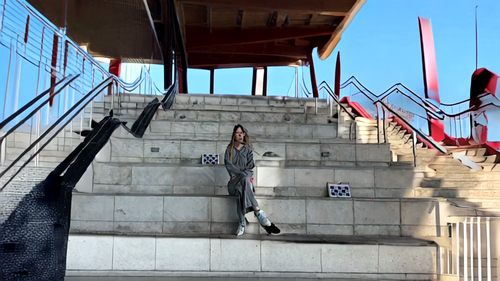} &
        \includegraphics[width=0.15\textwidth]{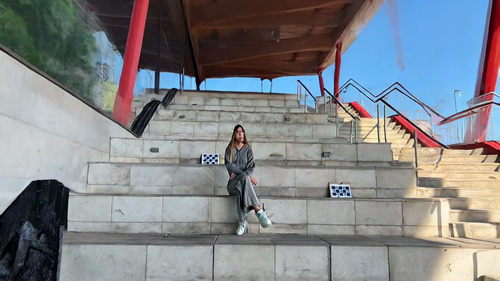} &
        \includegraphics[width=0.15\textwidth]{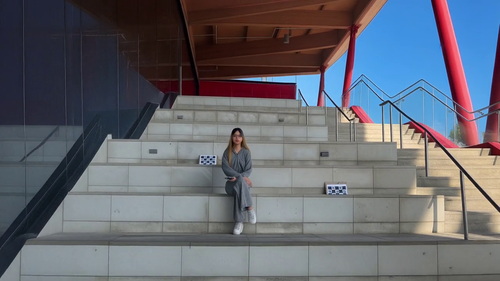} &
        \fbox{\includegraphics[width=0.15\textwidth]{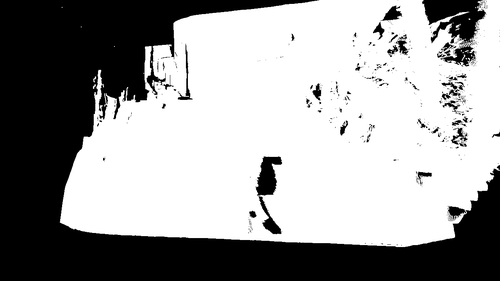}} &
        \includegraphics[width=0.15\textwidth]{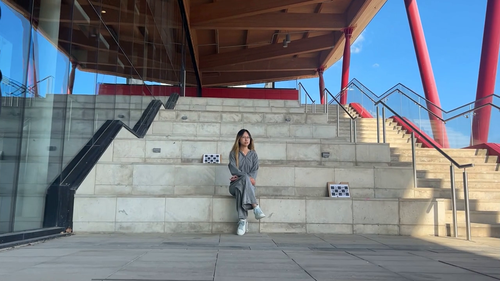} \\

        \includegraphics[width=0.15\textwidth]{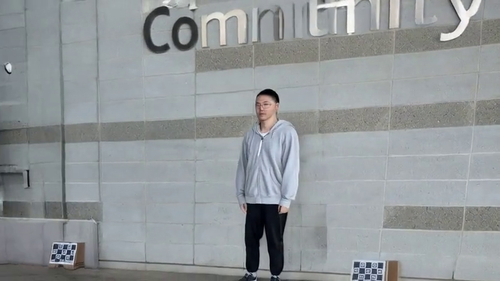} &
        \includegraphics[width=0.15\textwidth]{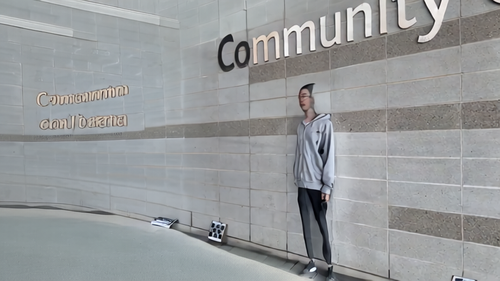} &
        \includegraphics[width=0.15\textwidth]{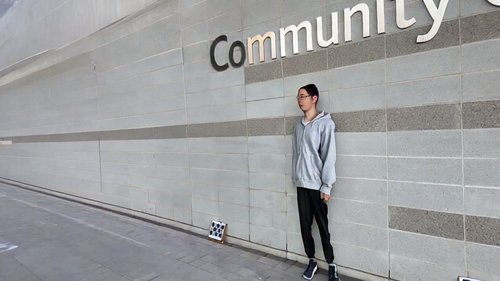} &
        \includegraphics[width=0.15\textwidth]{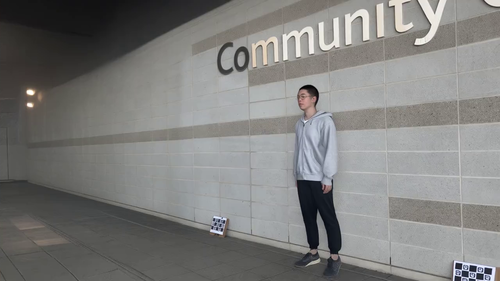} &
        \fbox{\includegraphics[width=0.15\textwidth]{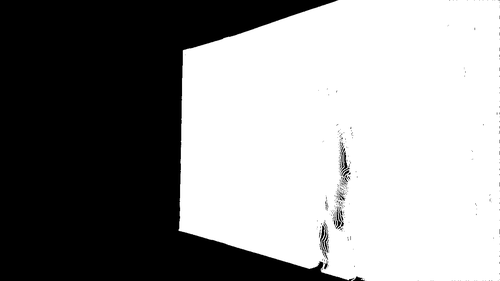}} &
        \includegraphics[width=0.15\textwidth]{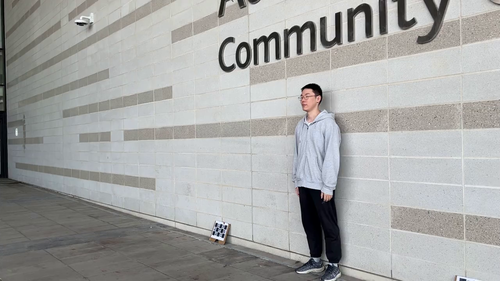} \\

        RecamMaster & TrajectoryCrafter & Gen3C & Ours & Visibility Mask & GT 
    \end{tabular}
    \caption{\textbf{More Qualitative comparison on the iPhone and iPhone-PTZ dataset}.}
    \vspace{5mm}
    \label{fig:supple_qualitative_iphone}
\end{figure*}

\begin{figure*}[t]
    \centering
    \setlength{\tabcolsep}{2pt}  %
    \renewcommand{\arraystretch}{1}  %
    \setlength{\colw}{2.9cm}
    \begin{tabular}{m{1cm}*{5}{>{\centering\arraybackslash}m{\colw}}}
        & 9s & 15s & 21s & 27s & $>$ 30s \\
        \centering Source &
        \includegraphics[width=\colw]{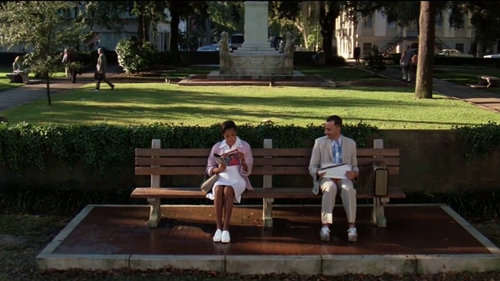} &
        \includegraphics[width=\colw]{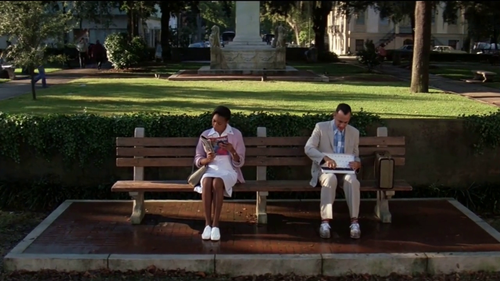} &
        \includegraphics[width=\colw]{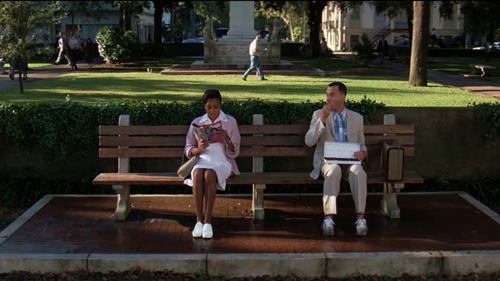} &
        \includegraphics[width=\colw]{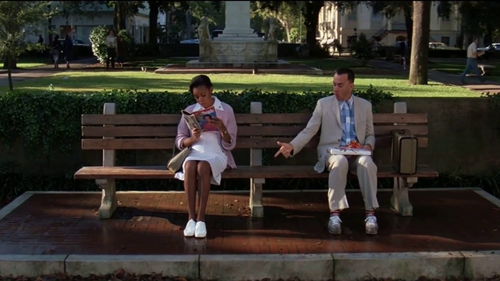} &
        \includegraphics[width=\colw]{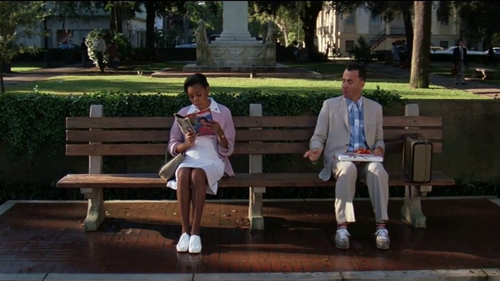} \\

        \centering Ours &
        \includegraphics[width=\colw]{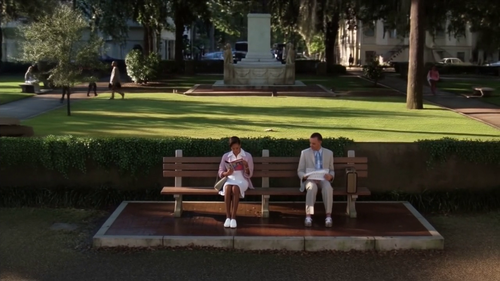} &
        \includegraphics[width=\colw]{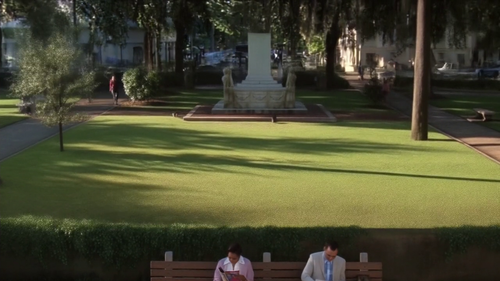} &
        \includegraphics[width=\colw]{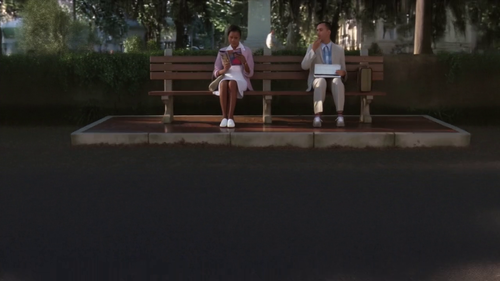} &
        \includegraphics[width=\colw]{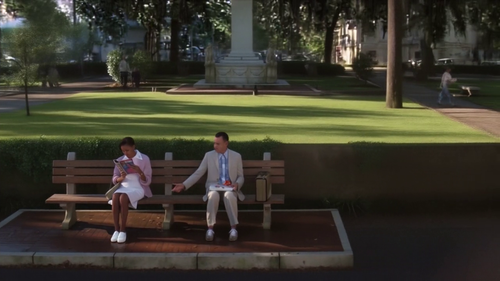} &
        \includegraphics[width=\colw]{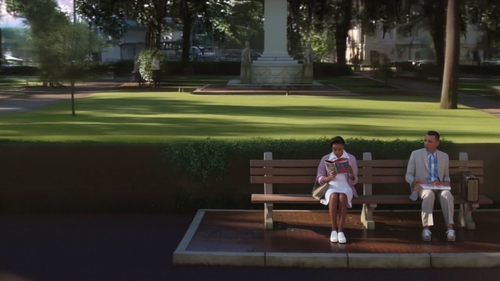} \\

        \centering Source &
        \includegraphics[width=\colw]{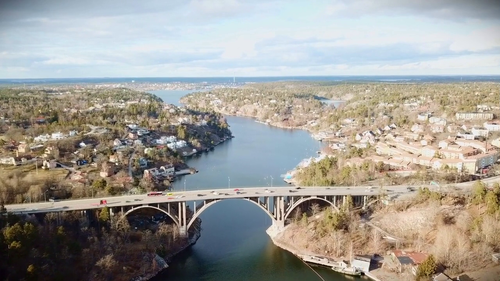} &
        \includegraphics[width=\colw]{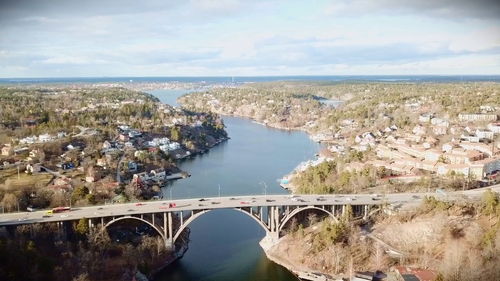} &
        \includegraphics[width=\colw]{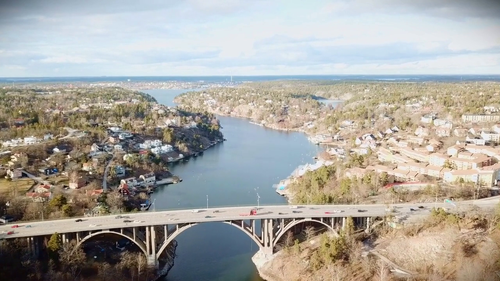} &
        \includegraphics[width=\colw]{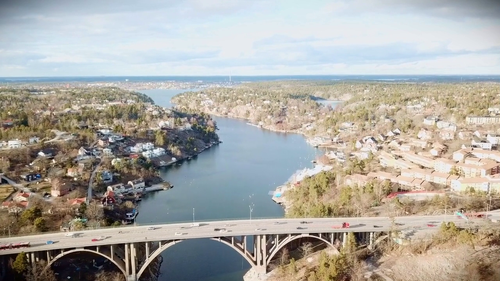} &
        \includegraphics[width=\colw]{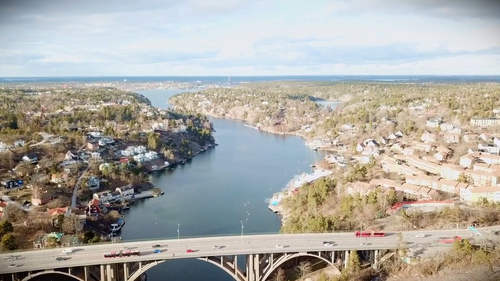} \\

        \centering Ours &
        \includegraphics[width=\colw]{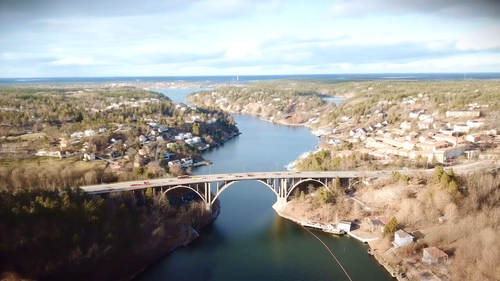} &
        \includegraphics[width=\colw]{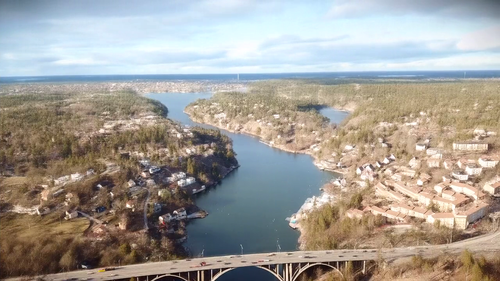} &
        \includegraphics[width=\colw]{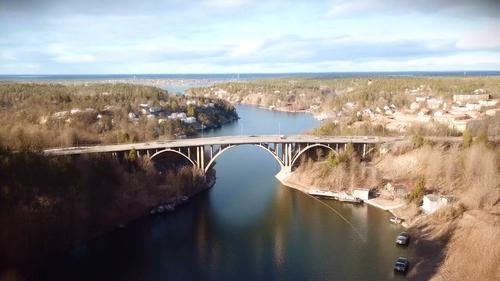} &
        \includegraphics[width=\colw]{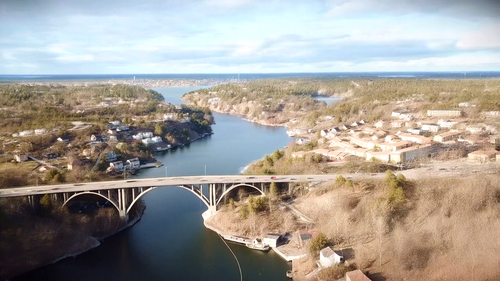} &
        \includegraphics[width=\colw]{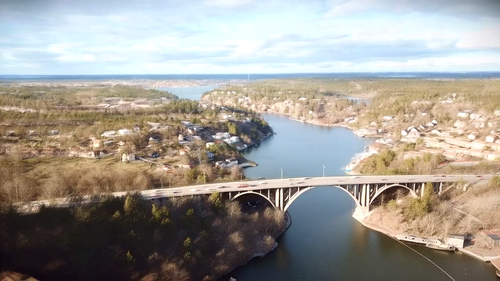} \\
        
    \end{tabular}
    \caption{\textbf{Qualitative results on in-the-wild long video generation}.}
    \label{fig:supple_qualitative_long_video}
\end{figure*}

More qualitative comparisons on the iPhone \cite{gao2022monocular} and iPhone-PTZ benchmarks are provided in Fig.~\ref{fig:supple_qualitative_iphone}.
Additional results on ultra–long video trajectory editing are shown in Fig.~\ref{fig:supple_qualitative_long_video}.
A \textbf{demo video} is also included to showcase the overall visual quality of our method.

\vspace{40mm}

\end{document}